\documentclass[10pt]{article}

\usepackage{nebius}

\usepackage{amsmath,amsfonts,bm}

\def\eqref#1{equation~\ref{#1}}

\def\1{\bm{1}}

\DeclareMathAlphabet{\mathsfit}{\encodingdefault}{\sfdefault}{m}{sl}
\SetMathAlphabet{\mathsfit}{bold}{\encodingdefault}{\sfdefault}{bx}{n}

\newcommand{\R}{\mathbb{R}}

\usepackage{amsmath}
\usepackage{tabularx}
\usepackage{colortbl}
\usepackage{tikz}
\usepackage{pgfplots}
\pgfplotsset{compat=1.18}
\usepgfplotslibrary{groupplots}
\usetikzlibrary{arrows.meta,calc,decorations.pathreplacing}
\usepackage{fontawesome5}
\usepackage{hyperref}
\nebiushypersetup
\hypersetup{
  pdftitle={Training Advisors for LLM Agents from Task Outcomes},
  pdfauthor={Sergei Polezhaev, Barys Liskavets, Ori Press, Alexander Golubev},
  pdfsubject={Preprint},
}

\definecolor{newblue}{RGB}{0,0,0}
\definecolor{revisionblue}{RGB}{0,70,190}

\definecolor{todored}{RGB}{170,25,35}

\newcommand{\methodname}{\textsc{Caddie}}

\providecommand{\nebiussans}{\sffamily}
\providecommand{\nebiussemibold}{\sffamily\fontseries{semibold}\selectfont}
\providecommand{\methodname}{\textsc{Caddie}}

\colorlet{caddiecolor}{nebiusdeepblue}
\colorlet{baselinecolor}{nebiuslavender}
\colorlet{improvedcolor}{nebiusdeepblue}
\colorlet{degradedcolor}{nebiuslavender}
\colorlet{nebiuscatone}{nebiusdeepblue}
\colorlet{nebiuscattwo}{nebiuslavender}
\colorlet{nebiuscatthree}{nebiuslime}
\colorlet{nebiuscatfour}{nebiusviolet}
\colorlet{nebiuscatfive}{nebiuslightblueii}
\colorlet{criticshade}{nebiuslightblueii}
\colorlet{analysisblue}{nebiusdeepblue}
\colorlet{analysisteal}{nebiusviolet}
\colorlet{analysisgold}{nebiuslimeii}
\colorlet{analysismauve}{nebiuslavender}
\colorlet{analysisfill}{nebiuslightblue}
\colorlet{neutralcolor}{nebiusdeepblue!55!white}
\colorlet{promptedmuted}{nebiuslavender}
\colorlet{kimimuted}{nebiuslime}
\colorlet{rrmmuted}{nebiuslightblueii}
\colorlet{figurebaseline}{baselinecolor}
\colorlet{figurecaddie}{caddiecolor}
\colorlet{figuregain}{improvedcolor}
\colorlet{figureharm}{degradedcolor}
\colorlet{gaincolor}{improvedcolor}
\colorlet{harmcolor}{degradedcolor}
\colorlet{nebiusmidgray}{nebiusdeepblue!45!white}
\colorlet{nebiusgray}{nebiusdeepblue!45!white}
\colorlet{nebiuslightgray}{nebiusdeepblue!12!white}
\colorlet{nebiushairline}{nebiusdeepblue!30!white}

\newcommand{\nebiusfont}{\nebiussans}

\pgfplotsset{
  nebius axis/.style={
    axis background/.style={fill=white},
    axis x line*=bottom,
    axis y line*=left,
    axis line style={draw=nebiusdeepblue!35,line width=0.4pt},
    y axis line style={draw=none},
    tick style={draw=nebiusdeepblue!35,line width=0.4pt},
    ytick style={draw=none},
    tick label style={font=\nebiusfont\scriptsize,text=nebiusdeepblue},
    x tick label style={align=center},
    label style={font=\nebiusfont\scriptsize,text=nebiusdeepblue},
    title style={font=\nebiussemibold\scriptsize,text=nebiusdeepblue},
    major tick length=2pt,
    ymajorgrids=true,
    grid style={draw=nebiusdeepblue!12,line width=0.4pt},
    legend style={font=\nebiusfont\scriptsize,draw=none,fill=none},
    every node near coord/.style={font=\nebiusfont\scriptsize,text=nebiusdeepblue,fill=none,inner sep=0pt},
    /pgf/number format/assume math mode=true,
  },
  nebius xbar axis/.style={
    nebius axis,
    ymajorgrids=false,
    xmajorgrids=true,
    y axis line style={draw=nebiusdeepblue!35,line width=0.4pt},
    x axis line style={draw=none},
    xtick style={draw=none},
    ytick style={draw=nebiusdeepblue!35,line width=0.4pt},
  },
  nebius value label/.style={
    nodes near coords={\pgfmathprintnumber[fixed,precision=1,assume math mode=true]{\pgfplotspointmeta}},
    every node near coord/.append style={yshift=1.5pt,anchor=south},
  },
  nebius baseline/.style={draw=none,fill=baselinecolor},
  nebius caddie/.style={draw=none,fill=caddiecolor},
  nebius single/.style={draw=none,fill=caddiecolor},
  nebius bar/.style={draw=white,line width=0.3pt,fill=#1},
  nebius pale bar/.style={draw=nebiushairline,line width=0.3pt,fill=#1},
}
\newcommand{\nebiuslegendswatch}[1]{%
  \raisebox{-0.12ex}{\tikz\fill[#1,rounded corners=0.8pt] (0,0) rectangle (9pt,1.45ex);}%
}
\newcommand{\nebiuslegendswatchsmall}[1]{%
  \raisebox{-0.12ex}{\tikz\fill[#1,rounded corners=0.6pt] (0,0) rectangle (7.5pt,1.45ex);}%
}
\newcommand{\nebiuslegendswatchpale}[1]{%
  \raisebox{-0.12ex}{\tikz\filldraw[fill=#1,draw=nebiushairline,line width=0.3pt,rounded corners=0.6pt] (0,0) rectangle (7.5pt,1.45ex);}%
}
\newcommand{\nebiussublabel}[1]{{\vspace{2pt}\nebiussemibold\footnotesize\color{nebiusdeepblue}#1\par}}
\newcommand{\nebiuslegendtext}{\nebiusfont\scriptsize\color{nebiusdeepblue}}

\tcbset{
  nebiusprompt/.style={
    breakable,
    enhanced,
    colback=nebiuslightblue,
    colframe=nebiuslightblue,
    boxrule=0pt,
    arc=2pt,
    left=7pt, right=7pt, top=5pt, bottom=5pt,
    borderline west={2pt}{0pt}{nebiusdeepblue},
  },
}

\begin{document}

\thispagestyle{empty}
\nebiuslogo

\begin{nebiustitlebox}
  \centering

  \vspace{0.3em}
  {\nebiustitlefont Training Advisors for LLM Agents from Task Outcomes\par}
  \vspace{1.1em}

  {\nebiusauthorfont
    \textbf{Sergei Polezhaev}\textsuperscript{1},\
    \textbf{Barys Liskavets}\textsuperscript{1},\
    \textbf{Ori Press}\textsuperscript{1},\
    \textbf{Alexander Golubev}\textsuperscript{1*}\par
  }
  \vspace{0.5em}

  {\nebiusaffilfont
    \textsuperscript{1}Nebius AI\\[0.3em]
    \textsuperscript{*}Correspondence to: alex\_golubev@nebius.com\par
  }
  \vspace{0.6em}

  \begin{abstract}

Large language model agents tackle multi-step tasks by interleaving reasoning and tool calls with observations from the environment. Prior work has shown that natural-language feedback can help these agents revise their decisions during task execution. We introduce \methodname{}, a method for training critics to provide natural-language analysis and advice as agents work through a task. Unlike approaches that rely on step-level labels or reference critiques, \methodname{} learns from whether the agent ultimately succeeds after receiving the critic’s feedback. We optimize the critic through reinforcement learning while keeping the base model frozen. Trained on multi-hop question answering with a single base model, our Qwen3-4B critic improves success rates across four base models of different scales and architectures, including three not used during critic training. On the MuSiQue benchmark, the trained critic improves Qwen3-4B's success rate by more than 25 percentage points, surpassing the performance of Kimi K3 without a critic. The same critic also yields gains on out-of-domain interactive benchmarks, including $\tau^3$ and DeepDive, with no additional training. Our results show that agents can decide when to seek help from a critic at inference time and that outcome-based critic training can produce guidance that transfers across base models and task domains.
\end{abstract}

\end{nebiustitlebox}

\section{Introduction}
\label{sec:introduction}

Large language model (LLM)-based agents \citep{react2022,taubench2024} increasingly tackle multi-step tasks by producing actions and reacting to environment observations. During its execution, an agent can get stuck repeating unsuccessful actions or lose sight of the original goal \citep{reflexion2023,reflact2025}. External feedback offers a way to help models reconsider their approach and steer the agent toward a successful solution by identifying problems and suggesting more useful next actions.

Outcome reward models (ORMs) \citep{cobbe2021} and process reward models (PRMs) \citep{uesato2022,lightman2023} address this need by assessing complete trajectories or intermediate steps, respectively. These models typically predict a score or verdict, sometimes preceded by textual reasoning, that can be used to rank candidates, identify faulty steps, or guide search. However, conventional step-level verification faces two practical limitations:
\begin{itemize}
    \item \textbf{Costly supervision.}
    Many PRMs require step-level training labels. Human annotation
    \citep{lightman2023} can provide these labels, but collecting them
    at scale is expensive \citep{mathshepherd2023}. Automatic labeling
    \citep{mathshepherd2023,automatedprocess2024} can also be costly
    and introduce noisy supervision.

    \item \textbf{Ambiguous step correctness.}
    Correctness-based assessment assumes that intermediate actions
    can be meaningfully classified as correct or faulty. In interactive environments, an apparently unproductive action may be needed to explore the environment or gather information. Recent agent PRMs \citep{agentprm2025} address this ambiguity by estimating future success, but using these scores to choose a better action still requires an additional selection or search procedure.
\end{itemize} 

Critique-based methods \citep{rl4f2023,ctrl2025,critiquerl2025} provide a complementary interface: natural-language feedback can explain a problem in the agent's current context and suggest how to proceed. By guiding the next action directly, such feedback can reduce the need for search, branching, or replay, which may be costly in some interactive environments. Such a method also makes potential errors and proposed next steps easier to interpret.

However, it is not straightforward how to train such critics effectively. Existing approaches train such critics through procedures that include synthesizing reference critiques (CTRL; \citealp{ctrl2025}) or separately optimizing their ability to judge and improve solutions (Critique-RL; \citealp{critiquerl2025}). These methods demonstrate the value of learned feedback and motivate simpler ways to train it for an agent that must continue acting and observing.

\begin{figure}[t]
  \centering
  \includegraphics[width=\linewidth]{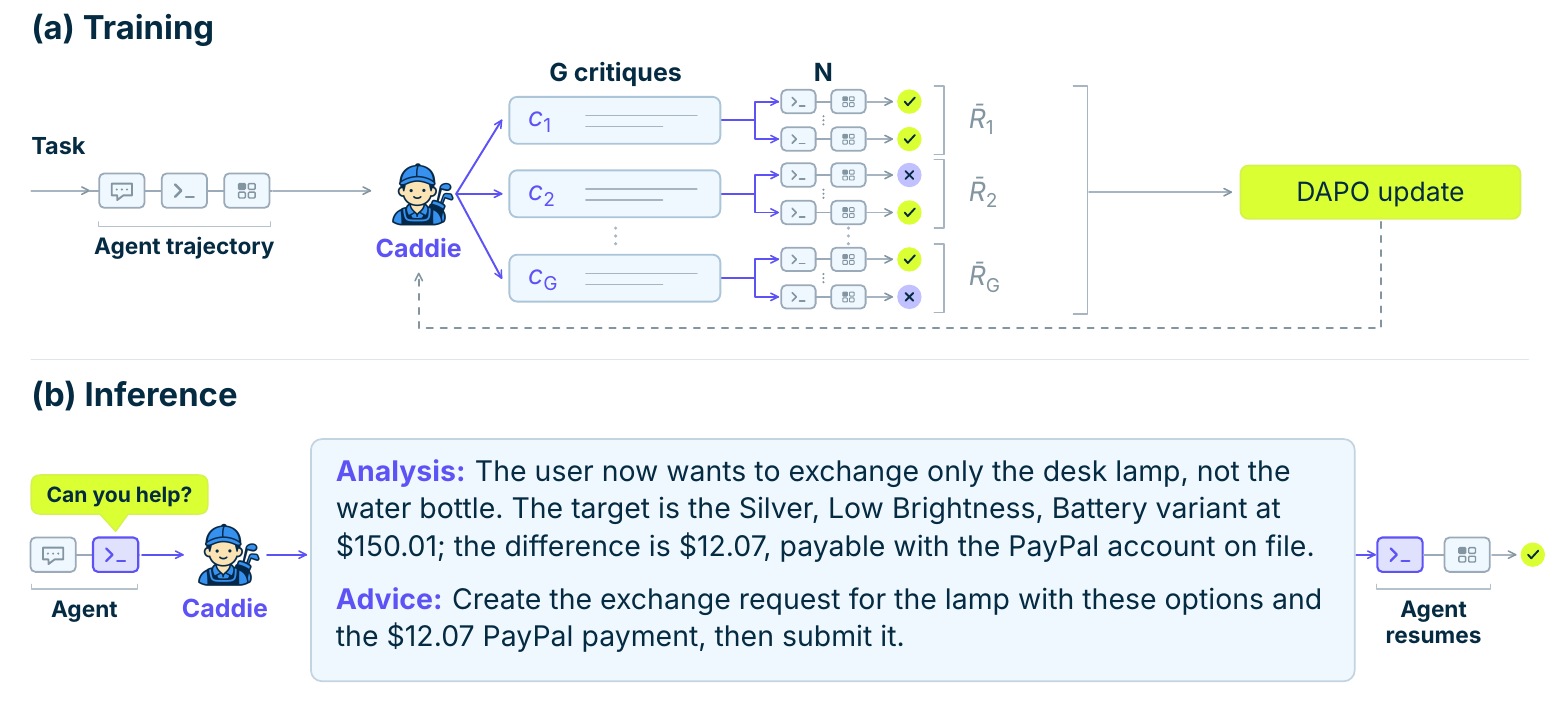}
  \caption{\textbf{Training and using \methodname{}.} (a)~Training: the frozen agent $\pi_b$ rolls out a task and a step $k$ is chosen randomly; at the prefix $s_k$ the critic $\pi_c^{\theta}$ samples $G$ critiques; each critique is then appended to the prefix and the frozen agent rolls out $N$ continuations until the end of the episode. The terminal rewards are normalized within the group and update only the critic with DAPO. (b)~Inference: the agent decides when to consult the critic through the \texttt{call\_critic} tool and receives the critique as the tool output.}
  \label{fig:caddie-overview}
\end{figure}

We introduce \methodname{}, a method for training critics to guide agents during task execution (Figure~\ref{fig:caddie-overview}). During training, we sample intermediate points in an agent's trajectory and ask the critic to provide natural-language analysis and advice. The agent then continues with this feedback, and we use its final task outcome as the reward for updating the critic through reinforcement learning, while keeping the agent frozen. We directly optimize the final outcome of the policy being steered, and require neither step-level annotations nor reference critiques. At inference time, the agent can decide when to consult the trained critic.

We evaluate our approach across multiple benchmarks and base models to study both in-domain effectiveness and generalizability to other base models and tasks. We also examine how the usefulness of feedback depends on the base model and the critic inference setup.

Our main contributions are:

\begin{itemize}
  \item \methodname: a method for training generative advisors to guide ongoing agent interactions using downstream task outcomes, without requiring step-level annotations or reference critiques.
  \item We show that a Qwen3-4B critic trained with one base model improves three additional models and yields gains on out-of-domain benchmarks without further training.
  \item We demonstrate useful feedback under various inference strategies, and analyze how its effectiveness differs across setups and base models.
\end{itemize}

\section{Method}
\label{sec:method}

We introduce \methodname{}, a method for training a generative critic that can provide natural-language feedback to an LLM agent at intermediate states within a trajectory. Section~\ref{sec:method:intervention} formalizes the interaction between the agent, the environment, and the critic, including when the critic is invoked. Section~\ref{sec:method:optimization} describes how we train the critic using the final outcomes of the base model's continuations while keeping the base model frozen.

\subsection{Critique intervention}
\label{sec:method:intervention}

\paragraph{Agent--environment interaction.}
For a task $x\sim\mathcal{D}$, the frozen base policy $\pi_b$ interacts with an environment $\mathcal{E}$. At step $t$, its state is defined as:
\begin{equation}
  s_t=(p,x,a_0,o_0,\ldots,a_{t-1},o_{t-1}),
  \label{eq:history}
\end{equation}
where $p$ is the system prompt, $a_i$ an action, and $o_i$ the corresponding observation. Each step samples an action and an environment response, then appends them to the state:
\begin{equation}
  a_t\sim\pi_b(\cdot\mid s_t),
  \qquad
  o_t\sim\mathcal{E}(\cdot\mid s_t,a_t),
  \qquad
  s_{t+1}=s_t\oplus(a_t,o_t).
  \label{eq:base-policy}
\end{equation}
 The episode ends at step $T$ with a terminating action or an exhausted interaction budget. The complete trajectory $\tau=s_{T+1}$ receives a final task reward $R(\tau)$.

\paragraph{Critic policy.}
The critic is a language model with policy $\pi_c^{\theta}$ and trainable parameters $\theta$. When invoked at step $t$, it observes state $s_t$ and generates a natural-language critique

\begin{equation}
  c_t\sim\pi_c^{\theta}(\cdot\mid s_t),
  \label{eq:critic-policy}
\end{equation}

consisting of an analysis of the state and advice for the next steps (the format is specified in Appendix~\ref{app:contract}). Then, the critique is appended as an additional message,

\begin{equation}
  \widetilde s_t=s_t\oplus c_t,
  \label{eq:augmented-history}
\end{equation}

and the agent samples its next action from the base policy conditioned on the augmented state, $a_t \sim \pi_b(\cdot \mid \widetilde{s}_t)$, following Eq.~\ref{eq:base-policy} for subsequent steps.
\paragraph{Intervention protocol.}
\label{sec:method:insertion}

An intervention protocol $\mathcal{P}$ determines whether to invoke the critic at each intermediate step $t$, i.e. $\mathcal{P}(s_t)\in\{0,1\}$. A trajectory may contain zero, one, or several critiques. We study three protocols:

\begin{enumerate}
  \item \textbf{Windowed intervention.} An intervention step $k$ is uniformly sampled from a predefined window $\mathcal{K}=\{k_{\min},\ldots,k_{\max}\}$. The critic is invoked if the episode reaches that step $k$, $\mathcal{P}_{\mathrm{win}}(s_t)=\mathbf{1}[t=k]$. The window can consist of a single step ($k_{\min}=k_{\max}$).
  \item \textbf{Probabilistic intervention.} At each step, the critic is invoked independently with a fixed probability $\alpha$, so $\mathcal{P}_{\mathrm{prob}}(s_t)\sim\mathrm{Bernoulli}(\alpha)$.
  \item \textbf{Self-call intervention.} The base model's action space includes an additional tool, $\texttt{call\_critic}$. The base policy decides when to consult the critic by selecting this tool, $\mathcal{P}_{\mathrm{self}}(s_t)=\mathbf{1}[a_t=\texttt{call\_critic}]$, and receives the critique as its output.
\end{enumerate}

Unless stated otherwise, we use $\mathcal{P}_{\mathrm{self}}$ for evaluation. The prompt and tool description are provided in Appendix~\ref{app:prompts}, and the protocols are compared in Appendix~\ref{app:policies}. Windowed and probabilistic intervention require no changes to the base model's prompt or tool definitions. However, they do not account for the current trajectory and may intervene too early, too late, or when no help is needed. Self-call intervention instead uses an adapted system prompt $\widetilde p$ describing the critic tool, while keeping the base model's parameters fixed. It lets the agent request help without an externally tuned schedule. Windowed or probabilistic intervention can perform better in some settings, but their performance depends on the intervention window or probability.

\subsection{Outcome-based critic training}
\label{sec:method:optimization}

We train the critic $\pi_c^\theta$ to generate feedback that improves the final reward of a frozen base model $\pi_b$.

\paragraph{Outcome-based rewards.}
We sample state prefixes $s_k$ using $\mathcal{P}_{\mathrm{win}}$ with $k \sim \mathrm{Uniform}(\mathcal{K})$, where $k_{\min}$ and $k_{\max}$ are fixed before training for each task environment. For each prefix, the critic $\pi_c^{\theta_{\mathrm{old}}}$ generates a group of $G$ critiques $c^{(1)},\ldots,c^{(G)}$. To evaluate each critique $c^{(i)}$, we append it to the prefix and run $N$ independent continuations, producing complete trajectories $\tau^{(i,n)} \sim P_b(\cdot \mid s_k \oplus c^{(i)})$. Here, $P_b$ denotes the trajectory distribution induced by the frozen base model and environment $\mathcal{E}$. We assign the critique the average final reward across these continuations,
\begin{equation}
  R_i = \frac{1}{N}\sum_{n=1}^{N} R\bigl(\tau^{(i,n)}\bigr).
  \label{eq:critique-reward}
\end{equation}
Thus, feedback is evaluated through its effect on downstream task performance.

\paragraph{Critic optimization.}
We use DAPO \citep{dapo2025}, which builds on GRPO \citep{deepseekmath2024} with asymmetric clipping, token-level loss aggregation, and dynamic sampling. Following GRPO, we compute the group-normalized advantage $\hat A_i=(R_i-\mu_R)/\sigma_R$, where $\mu_R$ and $\sigma_R$ are the mean and standard deviation of the rewards $R_1,\ldots,R_G$, and use the same advantage for every token in critique $c^{(i)}$. DAPO's dynamic sampling discards groups with zero reward variance and samples additional groups to fill the training batch. For a retained group, the clipped surrogate objective is

\begin{equation}
\begin{aligned}
\mathcal{J}_{\mathrm{DAPO}}(\theta)
={}&
\frac{1}{\sum_{i=1}^{G}|c^{(i)}|}
\sum_{i=1}^{G}\sum_{t=1}^{|c^{(i)}|}
\min\Bigl[
    \rho_{i,t}(\theta)\hat A_i, \
    \operatorname{clip}\!\bigl(
        \rho_{i,t}(\theta),
        1-\varepsilon_{\mathrm{low}},
        1+\varepsilon_{\mathrm{high}}
    \bigr)\hat A_i
\Bigr],
\end{aligned}
\label{eq:dapo}
\end{equation}

where $\rho_{i,t}(\theta)=\pi_c^\theta(c_t^{(i)}\mid s_k,c_{<t}^{(i)})/\pi_c^{\theta_{\mathrm{old}}}(c_t^{(i)}\mid s_k,c_{<t}^{(i)})$ is the token probability ratio. The objective averages over all generated critique tokens. Only the critic's parameters are updated; task instructions, observations, and base-model continuation tokens are excluded from the loss. Implementation details and hyperparameters are provided in Appendix~\ref{app:training_details}.

Other reward definitions are possible, such as improvement over a continuation without a critique; we discuss these alternatives in Appendix~\ref{app:reward_variants}. 

Such a formulation also constrains the choice of tasks and environments. For tasks solvable in a single step with a small token budget and no tool calls, an optimal critic may learn to solve the task itself and instruct the base model to copy its solution. We discuss this limitation further in Appendix~\ref{app:limitations}.

\section{Experiments}
\label{sec:experiments}

We evaluate whether outcome-based critic training improves task performance and whether the learned feedback transfers to base models and tasks unseen during critic training.

\subsection{Experimental setup}
\label{sec:experimental-setup}

\paragraph{Tasks and environments.}
We study in-domain performance on \textbf{MuSiQue} \citep{musique2021} and \textbf{ALFWorld} \citep{alfworld2020}, training a separate critic for each setting. MuSiQue requires combining evidence across two to four reasoning hops. We use two retrieval settings: \emph{attached}, in which the agent retrieves evidence from the 20 supplied passages, and \emph{wiki}, in which it searches Wikipedia. ALFWorld requires agents to navigate text-based household environments and interact with objects to complete tasks. To test transfer beyond the training tasks, we evaluate the MuSiQue-trained critics on \textbf{DeepDive} \citep{deepdive2025}, which requires web search to answer questions synthesized from knowledge graphs, and on \textbf{$\tau^3$} Retail and Airline \citep{taubench2024,tau3bench2026}, which require customer dialogue and tool use under domain-specific policies.

\paragraph{Critic training.}
All critics are initialized from Qwen3-4B-Instruct-2507 \citep{qwen34bmodelcard2025} and trained with DAPO as described in Section~\ref{sec:method:optimization}. We train separate critics for MuSiQue-attached, MuSiQue-wiki, and ALFWorld, keeping the Qwen3-4B base model frozen. For MuSiQue and ALFWorld, we use the authors' training split. Unless otherwise stated, we use $G=16$ critiques per state prefix and estimate each critique's reward using one base model continuation ($N=1$). Training runs use a single node with eight NVIDIA H100 GPUs, each with 80\,GB of memory. Training hyperparameters, intervention windows, and checkpoint selection are described in Appendix~\ref{app:training_details}.

\paragraph{Base models and transfer.}
We evaluate all three critics with Qwen3-4B, the base model used during training. On MuSiQue, we also evaluate the corresponding critics with Qwen3-30B-A3B-Instruct-2507 \citep{qwen330bmodelcard2025}, Qwen3.8-27B \citep{qwen38modelcard2026}, and Kimi K3 \citep{kimik3modelcard2026} to test how the critic generalizes to different base models. For transfer to DeepDive and $\tau^3$, we use Qwen3-4B and Qwen3.8-27B. The latter tests joint transfer across tasks and base models. For each trained variant, we use a single fixed critic checkpoint throughout evaluation. We abbreviate checkpoint names in tables.

\paragraph{Baselines.}
We compare \methodname{} with no critic, a prompted Qwen3-4B critic, and a prompted Kimi K3 critic. The prompted Qwen3-4B critic uses the same checkpoint as our critic's initialization, providing a fair comparison without outcome-based critic training. For transfer to unseen task families, we also include Agent-RRM \citep{agentrrm2026}, a generative reward model trained using teacher-generated reasoning, critiques, and trajectory scores. All main critic-assisted evaluations use the self-call protocol $\mathcal{P}_{\mathrm{self}}$, allowing the base model to request at most three critiques per trajectory.

\paragraph{Evaluation metric.}
For each condition, we generate eight independent rollouts per task using distinct seeds and fixed model checkpoints. We first average the binary success outcomes within each task, then report the mean of these per-task success rates across the evaluation set. The standard error of the mean (SEM) is the sample standard deviation of the per-task rates divided by the square root of the number of tasks. For the smaller DeepDive and $\tau^3$ evaluation sets (Table~\ref{tab:ood-main}), we instead compute the SEM across the eight seed-level aggregate accuracies, where each seed's success rate over all tasks is one sample.
\subsection{In-domain performance}
\label{sec:musique-results}
We evaluate each critic in its training task setting with Qwen3-4B as the base model. For MuSiQue, we use a random subset of 200 validation questions. We impose no step limit and use a maximum context length of 8,192 tokens. For ALFWorld, we use the official unseen validation split of 134 games, which are held out from the training split.

\begin{table}[htbp]
\centering
\small
\setlength{\tabcolsep}{4pt}
\renewcommand{\arraystretch}{1.12}
\caption{In-domain success rates (\%) with Qwen3-4B as the base model. Each \methodname{} critic is trained in the corresponding task setting. Entries report mean $\pm$ SEM across tasks, with eight rollouts per task. Bold marks the highest mean; italics mark other means whose $\pm$ SEM intervals overlap the leader's (not a significance test).}
\label{tab:main}
\label{tab:in-domain}
\begin{tabularx}{\linewidth}{Xrrr}
\toprule
 & \multicolumn{2}{c}{MuSiQue} & \\
\cmidrule(lr){2-3}
Critic & Attached & Wiki & ALFWorld \\
\midrule
No critic & $21.56 \pm 2.07$ & $10.38 \pm 1.59$ & $23.79 \pm 3.01$ \\
Prompted Qwen3-4B & $23.25 \pm 2.37$ & $16.06 \pm 2.19$ & $20.24 \pm 2.89$ \\
Prompted Kimi K3 & $29.13 \pm 2.46$ & $\textit{19.19} \pm 2.23$ & $30.22 \pm 2.80$ \\
\rowcolor{criticshade}
\methodname{} (ours) & $\mathbf{46.88} \pm 2.68$ & $\mathbf{21.13} \pm 2.38$ & $\mathbf{49.72} \pm 3.42$ \\
\bottomrule
\end{tabularx}
\end{table}

\paragraph{Outcome-based training improves feedback quality.}
\methodname{} achieves the highest mean success rate in both MuSiQue settings and on ALFWorld (Table~\ref{tab:in-domain}). Relative to the no-critic baseline, it improves success on MuSiQue by 25.31 percentage points with attached passages, 10.75 points with Wikipedia retrieval and by 25.93 points on ALFWorld. The trained 4B critic also outperforms both prompted critics in all three settings, indicating that outcome-based training produces more useful feedback than prompting alone in these evaluations.

\paragraph{Critic feedback after generator training.}
To study whether a critic can further improve a generator that has already been trained on the downstream task, we first trained the Qwen3-4B generator with DAPO-based optimization until it plateaus, then froze it and trained a critic through its continuations. On MuSiQue-attached, this additionally raises success from $49.69\%$ to $63.56\%$ ($+13.88$ percentage points), showing that critic feedback can provide further gains even after the generator itself has been trained on the task. This suggests that the two forms of training are complementary. Appendix~\ref{app:trained-generator} gives the training setup and results for both retrieval settings.

\subsection{Transfer to unseen base models}
\label{sec:base-transfer}
We pair the MuSiQue-trained critics with three base models unseen during critic training, retaining the retrieval settings and evaluation protocol above. This tests whether feedback learned through continuations from Qwen3-4B remains useful to other base models.

\begin{table}[htbp]
\centering
\small
\setlength{\tabcolsep}{4pt}
\renewcommand{\arraystretch}{1.12}
\caption{Transfer across base models on MuSiQue. Success rates (\%) are reported as mean $\pm$ SEM across tasks, with eight rollouts per task. \methodname{} critics are trained with Qwen3-4B and evaluated with each base model without further adaptation. Bold marks the highest mean; italics mark other means whose $\pm$ SEM intervals overlap the leader's (not a significance test).}
\label{tab:musique-main}
\label{tab:base-transfer}
\begin{tabularx}{\linewidth}{Xrrrr}
\toprule
 & & \multicolumn{2}{c}{Prompted critic} & \\
\cmidrule(lr){3-4}
Base model & No critic & Qwen3-4B & Kimi K3 & \methodname{} (ours) \\
\midrule
\multicolumn{5}{l}{\textbf{Attached passages}} \\
\addlinespace[2pt]
Qwen3-30B-A3B & $28.44 \pm 2.45$ & $34.75 \pm 2.82$ & $\textit{36.00} \pm 2.86$ & \cellcolor{criticshade}$\mathbf{41.63} \pm 2.83$ \\
Qwen3.8-27B & $46.63 \pm 2.93$ & $45.31 \pm 2.84$ & $41.13 \pm 2.77$ & \cellcolor{criticshade}$\mathbf{55.94} \pm 2.93$ \\
Kimi K3 & $38.56 \pm 2.87$ & $34.44 \pm 2.63$ & $30.19 \pm 2.36$ & \cellcolor{criticshade}$\mathbf{51.00} \pm 2.76$ \\
\midrule
\multicolumn{5}{l}{\textbf{Wikipedia retrieval}} \\
\addlinespace[2pt]
Qwen3-30B-A3B & $\textit{19.94} \pm 2.38$ & $\textit{20.50} \pm 2.43$ & $\textit{20.06} \pm 2.42$ & \cellcolor{criticshade}$\mathbf{20.75} \pm 2.41$ \\
Qwen3.8-27B & $\textit{26.69} \pm 2.69$ & $\textit{23.69} \pm 2.53$ & $21.44 \pm 2.36$ & \cellcolor{criticshade}$\mathbf{27.50} \pm 2.79$ \\
Kimi K3 & $20.06 \pm 2.29$ & $19.75 \pm 2.28$ & $18.75 \pm 2.18$ & \cellcolor{criticshade}$\mathbf{26.25} \pm 2.55$ \\
\bottomrule
\end{tabularx}
\end{table}

\paragraph{Feedback transfers across base models.}
\methodname{} improves over the no-critic baseline and achieves a higher mean than both prompted critics in all six conditions (Table~\ref{tab:base-transfer}). With attached passages, gains over the no-critic baseline range from 9.31 to 13.19 percentage points across the three models. Transfer is less consistent with Wikipedia retrieval, ranging from 0.81 to 6.19 points; Qwen3.8-27B increases from 26.69\% to 27.50\%. Thus, a critic trained with a 4B base model can benefit larger base models without adaptation, although the magnitude of the benefit depends on the base model and retrieval setting.

\subsection{Transfer to unseen task families}
\label{sec:ood-results}
We evaluate both MuSiQue-trained critics on 40 $\tau^3$ Retail tasks, 20 $\tau^3$ Airline tasks, and a subset of 64 DeepDive tasks without further critic training. In each critic-assisted condition, the frozen base model uses the self-call protocol to request feedback. For $\tau^3$, we use the updated benchmark environments with the predefined \texttt{test} splits inherited from $\tau^2$-bench. Evaluation with Qwen3-4B tests task transfer; evaluation with Qwen3.8-27B tests joint transfer across tasks and base models. For all out-of-domain evaluations, both the critic and the base model use a maximum context length of 32,768 tokens.

\begin{table}[htbp]
\centering
\small
\setlength{\tabcolsep}{4pt}
\renewcommand{\arraystretch}{1.12}
\caption{Transfer of MuSiQue-trained critics to unseen tasks. Entries report success rates (\%): the mean over all tasks and seeds $\pm$ the SEM across the eight seed-level aggregate accuracies. Shading marks our critics. Bold marks the highest mean; italics mark other means whose $\pm$ SEM intervals overlap the leader's (not a significance test).}
\label{tab:ood-main}
\label{tab:tau3-main}
\begin{tabularx}{\linewidth}{Xrrr}
\toprule
 & & \multicolumn{2}{c}{$\tau^3$} \\
\cmidrule(lr){3-4}
Critic & DeepDive & Retail & Airline \\
\midrule
\multicolumn{4}{l}{\textbf{Base model: Qwen3-4B}} \\
\addlinespace[2pt]
No critic & $26.95 \pm 2.38$ & $44.69 \pm 1.73$ & $31.25 \pm 1.57$ \\
Prompted Qwen3-4B critic & $28.91 \pm 1.39$ & $42.50 \pm 0.98$ & $31.88 \pm 2.49$ \\
Prompted Kimi K3 critic & $27.34 \pm 0.72$ & $40.31 \pm 1.53$ & $27.50 \pm 2.83$ \\
Agent-RRM & $29.30 \pm 0.75$ & $40.63 \pm 2.15$ & $\textit{33.75} \pm 2.63$ \\
\addlinespace[2pt]
\rowcolor{criticshade}
\methodname{} (attached) & $29.10 \pm 2.17$ & $45.00 \pm 1.34$ & $\mathbf{37.50} \pm 2.67$ \\
\rowcolor{criticshade}
\methodname{} (wiki) & $\mathbf{34.38} \pm 1.72$ & $\mathbf{48.75} \pm 1.25$ & $\textit{36.88} \pm 3.26$ \\
\midrule
\multicolumn{4}{l}{\textbf{Base model: Qwen3.8-27B}} \\
\addlinespace[2pt]
No critic & $\textit{48.83} \pm 1.61$ & $50.31 \pm 1.60$ & $\textit{74.38} \pm 1.75$ \\
Prompted Qwen3-4B critic & $\textit{46.09} \pm 3.60$ & $53.75 \pm 2.83$ & $\textit{74.38} \pm 2.58$ \\
Prompted Kimi K3 critic & $41.80 \pm 2.81$ & $\mathbf{59.69} \pm 2.73$ & $62.50 \pm 2.11$ \\
Agent-RRM & $\textit{48.44} \pm 3.54$ & $44.06 \pm 1.70$ & $49.38 \pm 5.13$ \\
\addlinespace[2pt]
\rowcolor{criticshade}
\methodname{} (attached) & $\textit{44.53} \pm 3.21$ & $\textit{59.06} \pm 1.70$ & $\textit{75.63} \pm 2.74$ \\
\rowcolor{criticshade}
\methodname{} (wiki) & $\mathbf{49.80} \pm 2.58$ & $\textit{58.12} \pm 2.44$ & $\mathbf{78.12} \pm 2.10$ \\
\bottomrule
\end{tabularx}
\end{table}

\paragraph{Feedback transfers beyond the training tasks.}
The attached-trained critic exceeds the no-critic mean in five of six base model--task combinations, and the wiki-trained critic in all six (Table~\ref{tab:ood-main}). With Qwen3.8-27B, the wiki-trained critic improves the mean on all three tasks, including $\tau^3$ Retail (50.31\% to 58.12\%) and Airline (74.38\% to 78.12\%). These results extend transfer beyond retrieval to customer-service interactions requiring dialogue and state-changing tool use.

Across the two trained variants, \methodname{} achieves the highest mean in five of six conditions; the prompted Kimi K3 critic leads on Qwen3.8-27B Retail. Transfer nevertheless depends on the training setting: on Qwen3.8-27B DeepDive, the attached-trained critic reduces success from 48.83\% to 44.53\%, whereas the wiki-trained critic yields a small increase to 49.80\%.

\section{Analysis}
\label{sec:analysis}

We study how the MuSiQue-wiki \methodname{} critic helps frozen generators solve problems outside its training domain.  The generator decides when to ask for advice, so the
analysis below describes the behavior of self-call rather than a randomized
intervention.  Additional breakdowns and qualitative examples appear in
Appendix~\ref{app:ood-selfcall-analysis}.

\paragraph{The generator usually asks once, and feedback helps when requested.}
Among rollouts that call \methodname{}, a single request is by far the most
common behavior.  A second request does not coincide with lower accuracy on
DeepDive.  The complete call-count distribution and the corresponding accuracy curves are shown in
Appendix~\ref{app:ood-selfcall-analysis}.  Figure~\ref{fig:ood-called-task-comparison}
compares feedback-bearing rollouts with no-critic rollouts on the same tasks.
The feedback-bearing group has a higher solve rate in all six settings, but
this comparison is conditional on the generator requesting feedback and
does not isolate its causal effect.

\begin{figure}[!ht]
  \centering
  \begin{tikzpicture}
\begin{axis}[
  nebius axis,
  width=\linewidth,
  height=4.9cm,
  ybar=3pt,
  bar width=16pt,
  ymin=0,
  ymax=90,
  ylabel={Solve rate (\%)},
  yticklabel={\pgfmathprintnumber[fixed,precision=0,assume math mode=true]{\tick}},
  symbolic x coords={Q4/DD,Q4/R,Q4/A,Q38/DD,Q38/R,Q38/A},
  xtick=data,
  xticklabels={{Q4\\DD},{Q4\\Retail},{Q4\\Airline},{Q38\\DD},{Q38\\Retail},{Q38\\Airline}},
  enlarge x limits=0.12,
  point meta=y,
  nebius value label,
  area legend,
  legend cell align={left},
  legend style={at={(0.015,0.97)},anchor=north west,legend columns=1,
    font=\nebiusfont\scriptsize,text=nebiusdeepblue,draw=none,fill=none},
  legend image post style={rounded corners=0.6pt},
]
\addplot+[nebius baseline] coordinates {
  (Q4/DD,20.000) (Q4/R,37.037) (Q4/A,29.167)
  (Q38/DD,44.907) (Q38/R,49.383) (Q38/A,66.667)
};
\addlegendentry{No critic}
\addplot+[nebius caddie] coordinates {
  (Q4/DD,30.357) (Q4/R,54.545) (Q4/A,50.000)
  (Q38/DD,57.258) (Q38/R,53.846) (Q38/A,75.862)
};
\addlegendentry{\methodname{} feedback}
\end{axis}
\end{tikzpicture}
  \caption{Solve rates on tasks where \methodname{} delivered feedback. Navy: feedback-bearing rollouts; lavender: no-critic rollouts on the same tasks. This generator-selected subset gives a descriptive comparison, not a causal estimate. Q4/Q38 denote Qwen3-4B/Qwen3.8-27B; DD denotes DeepDive.}
  \label{fig:ood-called-task-comparison}
\end{figure}
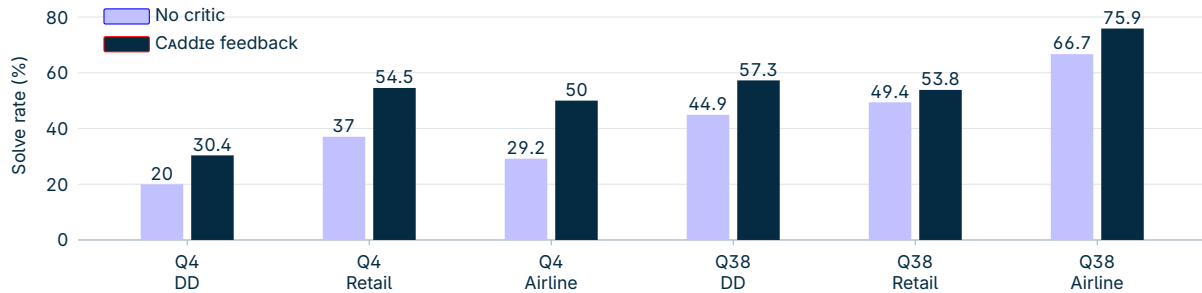

\begin{figure}[!hb]
\centering
\includegraphics[width=0.8\linewidth]{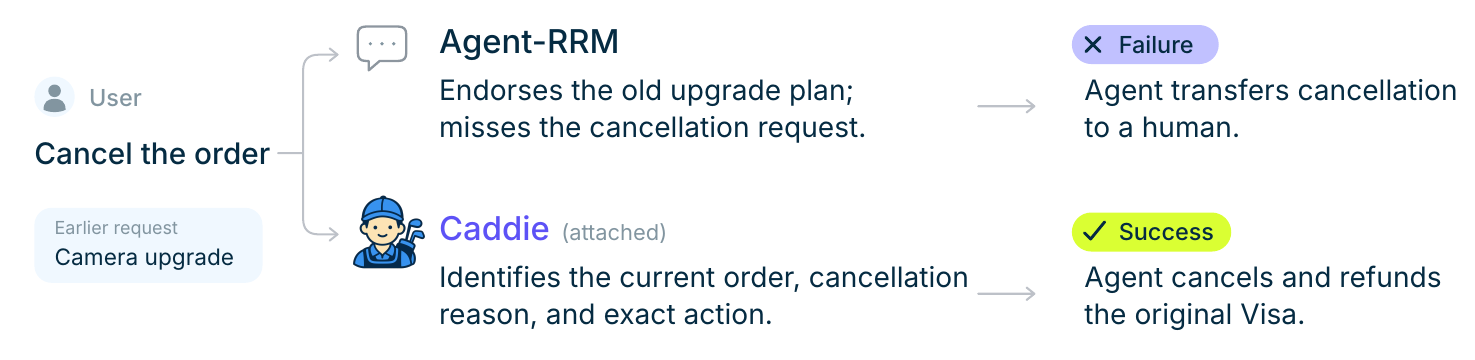}
\caption{Matched $\tau^3$ Retail trajectories after the same
change in user intent.  Each path summarizes the critic advice and the
frozen-generator continuation.}
\label{fig:tau3-trajectory-main}
\end{figure}

\paragraph{The advice adapts to the task.}
On DeepDive, \methodname{} mainly suggests either another evidence search or
that the generator should stop and answer.  On $\tau^3$, particularly with Qwen3.8-27B, it more often
recommends verifying state, asking the user, or executing an action.  Thus the critic does
not repeat one generic instruction: its advice changes with the environment.
Figure~\ref{fig:tau3-trajectory-main} illustrates this behavior after a $\tau^3$
user changes intent.  Additional distributions and examples are provided in
Appendix~\ref{app:ood-selfcall-analysis}.

\begin{figure}[!ht]
  \centering
  \begin{minipage}[t]{0.49\linewidth}
\centering
\begin{tikzpicture}
\begin{axis}[
  nebius axis,
  width=\linewidth,
  height=3.9cm,
  ybar=2pt,
  bar width=16pt,
  ymin=0,
  ymax=92,
  ylabel={Solve rate (\%)},
  yticklabel={\pgfmathprintnumber[fixed,precision=0,assume math mode=true]{\tick}},
  symbolic x coords={Answer,Search},
  xtick=data,
  enlarge x limits=0.45,
  point meta=y,
  nebius value label,
]
\addplot+[nebius baseline] coordinates {(Answer,25.000) (Search,19.318)};
\addplot+[nebius caddie] coordinates {(Answer,29.412) (Search,30.526)};
\end{axis}
\end{tikzpicture}

\nebiussublabel{(a) Q4 / DeepDive}
\end{minipage}\hfill
\begin{minipage}[t]{0.49\linewidth}
\centering
\begin{tikzpicture}
\begin{axis}[
  nebius axis,
  width=\linewidth,
  height=3.9cm,
  ybar=2pt,
  bar width=16pt,
  ymin=0,
  ymax=92,
  yticklabel={\pgfmathprintnumber[fixed,precision=0,assume math mode=true]{\tick}},
  symbolic x coords={Answer,Search},
  xtick=data,
  enlarge x limits=0.45,
  point meta=y,
  nebius value label,
]
\addplot+[nebius baseline] coordinates {(Answer,51.471) (Search,36.806)};
\addplot+[nebius caddie] coordinates {(Answer,77.941) (Search,30.909)};
\end{axis}
\end{tikzpicture}

\nebiussublabel{(b) Q38 / DeepDive}
\end{minipage}

\vspace{1.5mm}
\centering
{\nebiuslegendtext
\nebiuslegendswatch{baselinecolor}\,Retained baseline\qquad
\nebiuslegendswatch{caddiecolor}\,\methodname{} with feedback}
  \caption{DeepDive solve rate by advice type. Lavender (retained baseline): no-critic rollouts on the tasks where that advice type was given; navy: rollouts that received it.}
  \label{fig:deepdive-advice-by-generator}
\end{figure}
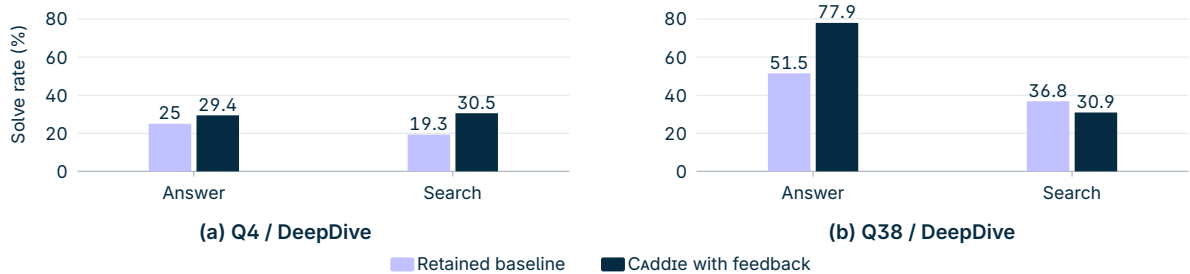

\paragraph{The most useful advice depends on generator scale.}
Figure~\ref{fig:deepdive-advice-by-generator} shows a simple contrast.  The
smaller Qwen3-4B generator benefits most from targeted search advice, matching
its tendency to stop before collecting enough evidence.  The larger
Qwen3.8-27B generator benefits most from being told to answer with evidence it
already has, matching its tendency to keep searching after the answer is
supported.  Additional search can therefore help the smaller generator but
distract the larger one.

Together, these results suggest a simple OOD mechanism: the generator calls
\methodname{} selectively, and the critic adapts its correction to both the
target domain and the generator's characteristic weakness.

\section{Discussion and Limitations}
\label{sec:discussion}

\paragraph{Task structure.}
Our method is intended for multi-step, tool-using tasks whose solutions require further environment interactions. On self-contained tasks, such as step-by-step mathematical reasoning, the critic can instead learn to generate complete solutions for the base model to copy, which can be optimal under the outcome-based objective. In this case, the critic learns to act as a task-solving policy, limiting its ability to transfer across base models and tasks. We discuss this further in Appendix~\ref{app:limitations}.

\paragraph{Practical constraints.}
Training requires $GN$ base-model continuations per sampled state and the ability to restore the same environment state across branches. This can be expensive for long trajectories, large base models, or costly tool calls. We train only Qwen3-4B critics, leaving the effect of critic scale on performance and transfer unverified.

\paragraph{Scalar verification.}
Our formulation treats critique as an intervention rather than a correctness judgment about an individual step. Scalar verifiers can rank or select candidate continuations, whereas a generative critic supplies text that conditions the base model's next actions. These approaches can be combined. Our experiments do not establish that natural-language feedback is generally superior to scalar verification; such a comparison would require matched inference procedures and compute budgets.

\section{Related Work}
\label{sec:related_work}
\paragraph{Learning critics from task outcomes.}
RL4F \citep{rl4f2023} and CTRL \citep{ctrl2025} train critics to improve a frozen model's revised outputs. Both methods initialize the critic with supervised critique examples: RL4F uses human-written or programmatically generated feedback, while CTRL synthesizes critiques using code-execution feedback. Critique-RL \citep{critiquerl2025} first trains correctness judgments, then optimizes feedback for successful revisions while preserving judgment accuracy. We train the critic directly from the final outcomes of agent continuations, providing advice during an ongoing trajectory without supervised critique initialization or a separate correctness-judgment stage.

\paragraph{Learned feedback for agents.}
Steer, Don't Solve \citep{steer2026} trains critics using teacher feedback and judge-ranked preferences. Agent-RRM \citep{agentrrm2026} trains an evaluator whose critiques guide revisions of completed attempts. Closest to our work, Advisor Models \citep{advisor2026} train a small advisor with GRPO to write per-instance advice for a frozen student, rewarded only by the student's task reward, and show that an advisor trained with a low-cost student transfers to frontier models. We share this training signal but differ in where credit is assigned, when advice arrives, and what transfers. They score advice by the student's response to the whole query, delivered before the response, after a completed attempt, or every five steps in their coding-agent experiments; we branch from a sampled intermediate state and compare $G$ critiques through continuations from that same state, so the critic learns to correct an episode in progress rather than to prompt a fresh attempt. Their advice follows a fixed schedule; our agent requests it through \texttt{call\_critic} when uncertain, avoiding the schedule sensitivity shown in Appendix~\ref{app:policies}. They report transfer across students and no degradation on other benchmarks; we show gains for unseen base models on unseen task families without any adaptation.

\paragraph{Revision and reflection.}
\citet{selfcorrect2023} train a separate corrector that directly rewrites a generator's outputs. GLoRe \citep{glore2024} similarly trains models to revise solutions, using reward models to locate errors and guide global or local revisions. Retroformer \citep{retroformer2023} learns reflections on failed trajectories to guide subsequent attempts. We instead train the critic to advise the agent on its next steps during the same attempt.

\paragraph{Training agents with feedback.}
ECHO \citep{echo2026} trains the critic alongside the agent so that feedback adapts as the agent learns. PivoARL \citep{pivoarl2026} learns both reflection and action generation from retries after failures. ICRL \citep{icrl2026} trains a shared model as both solver and critic, while \citet{taiscore2026} alternate critic and generator training for open-ended generation. Our training updates only the critic, leaving the acting model's parameters unchanged.

\section{Conclusion and Future Work}
\label{sec:conclusion}

We introduced \methodname{}, a method for training natural-language critics to provide analysis and advice during an agent's interaction. We adapt the GRPO-based optimization generally used for policy training to learn critiques from the final outcomes of a frozen base model's continuations. Using task outcomes as a common training signal avoids the need for task-specific step labels or reference critiques.

On MuSiQue, our trained Qwen3-4B critic improves the Qwen3-4B base model's success rate by 25.31 percentage points in the attached setting. The critic also transfers to base models not used during training, including Qwen3-30B-A3B, Qwen3.8-27B, and Kimi K3, and yields gains on out-of-domain tasks in $\tau^3$ and DeepDive without further training. These results show that the same training procedure can produce feedback that is useful across base models and task settings.

\begin{samepage}
We also identify a limitation concerning task structure. In self-contained tasks that can be solved in one pass, the critic can learn to generate complete solutions for the base model to copy, which can be optimal under the outcome-based objective.

Future work could explore fine-tuning the base policy to use a trained critic more effectively and explore joint training of the base policy and critic.
\par
\end{samepage}

\bibliography{submission}
\bibliographystyle{nebius}

\appendix

\clearpage
\section{Alternative Objectives}
\label{app:reward_variants}
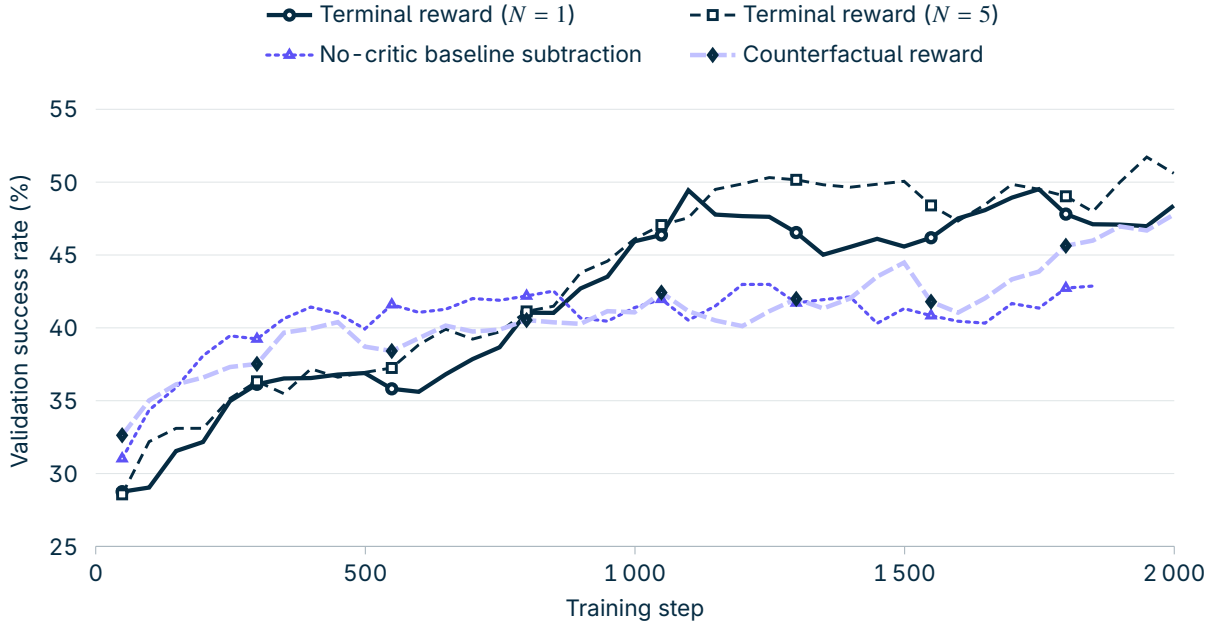
\begin{figure}[htbp]
\centering
\begingroup
\def\RewardPlotMaxStep{2000}
\def\RewardPlotSmoothingWeight{0.5}
\colorlet{curveone}{nebiusdeepblue}
\colorlet{curvetwo}{nebiusdeepblue}
\colorlet{curvethree}{nebiusviolet}
\colorlet{curvefour}{nebiuslavender}

\begin{tikzpicture}
\begin{axis}[
  nebius axis,
  width=0.96\linewidth, height=215pt,
  xmin=0, xmax=\RewardPlotMaxStep, ymin=25, ymax=56,
  restrict x to domain=0:\RewardPlotMaxStep,
  xtick={0,500,1000,1500,2000}, ytick={25,30,35,40,45,50,55},
  xlabel={Training step}, ylabel={Validation success rate (\%)},
  label style={font=\nebiusfont\small,text=nebiusdeepblue},
  tick label style={font=\nebiusfont\small,text=nebiusdeepblue},
  scaled ticks=false,
  xticklabel style={/pgf/number format/1000 sep={\,}},
  tick align=outside,
  major tick length=3pt,
  unbounded coords=jump,
  y filter/.code={
    \edef\RewardPlotCurrentY{#1}
    \def\RewardPlotMissingY{nan}
    \ifx\RewardPlotCurrentY\RewardPlotMissingY
      \def\pgfmathresult{nan}
    \else
      \ifnum\coordindex=0
        \xdef\RewardPlotPreviousY{#1}
      \fi
      \pgfmathparse{\RewardPlotSmoothingWeight*(#1)+(1-\RewardPlotSmoothingWeight)*\RewardPlotPreviousY}
      \xdef\RewardPlotPreviousY{\pgfmathresult}
    \fi
  },
  legend columns=2,
  legend style={
    at={(0.5,1.04)}, anchor=south, draw=none, fill=none,
    font=\nebiusfont\small, text=nebiusdeepblue, cells={anchor=west},
    /tikz/every even column/.append style={column sep=15pt},
    /tikz/row sep=3pt,
  },
  every axis plot/.append style={line width=1.2pt, mark size=1.7pt, mark repeat=5},
  clip=true,
]
\addplot[color=curveone, solid, line width=1.6pt, mark=*, mark options={solid,fill=white}]
  table[x=step,y=terminal_n1_pct,col sep=comma]{figures/data/reward_variants_validation.csv};
\addlegendentry{Terminal reward ($N=1$)}

\addplot[color=curvetwo, dash pattern=on 4pt off 2.5pt, line width=1.1pt, mark=square*, mark options={solid,fill=white}]
  table[x=step,y=terminal_n5_pct,col sep=comma]{figures/data/reward_variants_validation.csv};
\addlegendentry{Terminal reward ($N=5$)}

\addplot[color=curvethree, dotted, line cap=round, line width=1.2pt, mark=triangle*, mark options={solid,fill=white}]
  table[x=step,y=baseline_subtraction_pct,col sep=comma]{figures/data/reward_variants_validation.csv};
\addlegendentry{No-critic baseline subtraction}

\addplot[color=curvefour, dash pattern=on 6pt off 1.8pt, line width=1.7pt, mark=diamond*, mark options={solid,fill=nebiuslavender,draw=nebiusdeepblue}]
  table[x=step,y=gumbel_counterfactual_pct,col sep=comma]{figures/data/reward_variants_validation.csv};
\addlegendentry{Counterfactual reward}
\end{axis}
\end{tikzpicture}
\endgroup
\caption{MuSiQue validation success on a 20-task subset for four reward variants, shown up to 2,000 training steps. All curves use an exponential moving average with weight $0.5$ on the current evaluation. Confidence bands are unavailable; these smoothed traces illustrate training dynamics but do not establish differences in performance, convergence speed, or reward variance.}
\label{fig:reward-variants-validation}
\end{figure}

Alongside the terminal-outcome reward used in the main method, we explore alternative reward definitions and continuation counts. Figure~\ref{fig:reward-variants-validation} compares MuSiQue validation curves on a subset of 20 tasks with $N=1$ and $N=5$, alongside baseline subtraction and a Gumbel counterfactual reward.

\paragraph{Terminal outcome.}
For the outcome-trained critics, $R_i=\frac{1}{N}\sum_{n=1}^N R(\tau^{(i,n)})$, where $R(\tau^{(i,n)})\in\{0,1\}$ indicates final task success. The main experiments use $N=1$; we also evaluate $N=5$. Figure~\ref{fig:reward-variants-validation} shows broadly similar validation curves, with smaller fluctuations for $N=5$ after roughly 1,000 training steps. Final performance on the full validation set produces similar results for both variants. Since $N=1$ requires fewer base-model continuations and therefore less compute, we use it in the main experiments.

\paragraph{Gumbel-coupled counterfactual reward.}
We explore a paired reward comparison inspired by Gumbel-Max structural causal models \citep{oberst2019counterfactual}. For each critique $c^{(i)}$ at state $s_k$, we run two continuations with the frozen base model, one with the critique and one without it:
\[
\tau_i^{\mathrm{crit}} = F_b(s_k\oplus c^{(i)};\xi_i),
\qquad
\tau_i^{0} = F_b(s_k;\xi_i).
\]
Here, $F_b$ denotes a rollout of the base model and environment, and $\xi_i$ supplies shared sampling noise. Both continuations start from the same environment state. Under Gumbel-max sampling, each generated token is selected as
\[
z_j=\arg\max_v\left(\log q_j(v)+g_{j,v}\right),
\qquad
g_{j,v}\overset{\mathrm{iid}}{\sim}\mathrm{Gumbel}(0,1),
\]
where $q_j(v)$ is the base model's probability of token $v$ given the current context in that continuation. The two continuations reuse the same noise $g_{j,v}$ at corresponding generation positions $j$, while their contexts may differ. The critic receives the difference in final task rewards,
\[
R_i^{\mathrm{CF}}=R(\tau_i^{\mathrm{crit}})-R(\tau_i^{0}).
\]
The coupling is positional, not semantic. The contexts differ in length as soon as the critique is inserted, and matching generation positions need not correspond to the same action or reasoning step once the continuations diverge. Shared noise therefore does not guarantee matching tokens, tool calls, or environment observations. It reduces the variance of the reward difference relative to independent sampling only when the paired rewards have positive covariance. We do not measure how much correlation survives trajectory divergence, so we cannot quantify the resulting variance reduction.

The plotted counterfactual run rises earlier, but the small validation subset and absence of confidence bands do not establish faster convergence. Since this reward also requires no-critique continuations, we retain the simpler terminal-outcome reward for the main experiments.

\paragraph{No-critique baseline subtraction.}
For each state $s_k$, we sample $M$ independent continuations from the frozen base model without a critique and estimate its expected final reward as
\[
\widehat b_0(s_k)
=\frac{1}{M}\sum_{m=1}^{M} R(\tau^{0,m}),
\qquad
\tau^{0,m}\sim P_b(\cdot\mid s_k).
\]
Instead of centering critique rewards by their group mean, we use the advantage
\[
\widehat A_i^{0}=R_i-\widehat b_0(s_k).
\]
Critiques whose estimated reward exceeds the no-critique baseline receive positive advantage; those below it receive negative advantage.

Because the baseline is sampled independently of the critique and treated as constant during optimization, its contribution to the expected, unclipped score-function gradient is zero:
\[
\mathbb E_{c\sim\pi_c^\theta(\cdot\mid s_k)}\!\left[
\widehat b_0(s_k)\nabla_\theta\log\pi_c^\theta(c\mid s_k)
\right]=0.
\]
This identity does not imply that changing the baseline leaves the clipped training objective unchanged. The baseline may reduce gradient variance by accounting for differences in state difficulty, but its estimation introduces additional noise. For independent continuations, this noise decreases with $M$:
\[
\operatorname{Var}\!\left[\widehat b_0(s_k)\mid s_k\right]
=\frac{\operatorname{Var}_{\tau\sim P_b(\cdot\mid s_k)}[R(\tau)]}{M}.
\]

In the plotted run, this variant improves early and later falls below the terminal-outcome curves; without uncertainty estimates, however, this does not establish a reliable ranking. One possible explanation concerns the signs of the advantages. Early in training, the no-critique baseline assigns negative advantages to feedback that worsens task outcomes. Later, if most critiques outperform this baseline, even critiques that are weaker than others in the group can receive positive advantages. This may affect optimization under clipping. A possible extension is to gradually shift from the no-critique baseline to the group-mean baseline during training; we leave this for future work.

\paragraph{Self-call during training.}
The main experiments use windowed intervention $\mathcal{P}_{\mathrm{win}}$ during training. We also explore training with the self-call protocol $\mathcal{P}_{\mathrm{self}}$. This variant achieves similar in-domain performance but transfers less consistently to other base models and tasks. One possible explanation is that different base models request feedback at different states. Training only on states where one base model chooses to call the critic may therefore limit its exposure to states encountered with other base models.

\section{Format}
\label{app:contract}

\paragraph{Feedback format.}
The critic is instructed to analyze the trajectory and advise the base model on its next steps without calling tools, completing the task, or providing a complete action sequence. The two required sections are:
\begin{quote}\small\ttfamily
Analysis: <brief diagnosis>\par
Advices for next steps: <concise guidance>
\end{quote}
Both sections must be nonempty. During training, malformed feedback receives a reward of $-1$ without running a base-model continuation.

\paragraph{Tool semantics and message placement.}
In self-call evaluation, the base model emits brief reasoning followed by the empty tag \texttt{<critic/>}, which implements the \texttt{call\_critic} action of Section~\ref{sec:method:intervention}. Feedback is delivered in a \texttt{user}-role observation message with the prefix \texttt{Critic feedback:}; the full preceding base-model conversation is retained. A critic call does not advance the environment or consume an environment action. The main evaluations allow at most three critic calls per trajectory; after the third, the observation explicitly permits only an environment action.

\paragraph{Budgets and termination.}
ALFWorld evaluation allows 50 environment actions, an 8,192-token context, 128 output tokens per base-model decision, and 512 tokens per critique. Training uses a 12-action budget, shared by the prefix and continuation (Appendix~\ref{app:training_details}). An episode ends on success, exhaustion of the action budget, malformed output, an invalid tool call, or context overflow.

\section{Tasks, Splits, and Evaluators}
\label{app:tasks}

\paragraph{MuSiQue.}
In the attached setting, the base model searches the 20 passages supplied with each MuSiQue question \citep{musique2021}. In the wiki setting, it queries a frozen Wikipedia retrieval service. The main comparisons use the same 200 validation questions and eight trials per question (1,600 trajectories per condition). This subset contains 104 two-hop, 63 three-hop, and 33 four-hop questions.

The attached-setting results with trained critics as well as all the no-critic or prompted critic baselines in the main tables require at least three searches. The evaluator checks for a nonempty normalized exact match against any reference answer. Normalization consists of case-folding, replacing punctuation with spaces, removing English articles, and collapsing whitespace.

\paragraph{ALFWorld.}
We use the ALFWorld text environment \citep{alfworld2020} and the standard \texttt{json\_2.1.1} data. The official splits contain 3,553 training games, 140 seen-validation games, and 134 unseen-validation games. We restrict critic training to place, heat, and examine-under-light tasks, yielding 1,557 eligible games (790 place, 459 heat, and 308 examine-under-light). Evaluation covers all six task families on all 134 unseen-validation games, with eight trials per game.

\paragraph{DeepDive and $\tau^3$.}
The transfer evaluations use 64 DeepDive questions \citep{deepdive2025}, 40 $\tau^3$ Retail tasks, and 20 $\tau^3$ Airline tasks \citep{tau3bench2026}, with eight trials per task and a 32,768-token context. The $\tau^3$ evaluations use the predefined \texttt{test} splits inherited from $\tau^2$-bench. For DeepDive, the base model searches through Tavily, and a Qwen3-32B judge evaluates final answers in non-thinking mode.

\section{Models and Training Details}
\label{app:training}
\label{app:training_details}

\paragraph{Model roles.}
In the main experiments, the critic is initialized from Qwen3-4B-Instruct-2507, and all its parameters are trainable. The Qwen3-4B-Instruct-2507 base model remains frozen.

\begin{table}[htbp]
\centering\small
\caption{Optimization settings for MuSiQue-attached, MuSiQue-wiki, and ALFWorld.}
\label{tab:training-config}
\begin{tabularx}{\linewidth}{lX}
\toprule
Parameter & Value \\
\midrule
Objective & Token-level DAPO; lower/upper clip $0.20/0.28$ \\
Optimizer & AdamW; learning rate $10^{-6}$; $\beta_1=0.9$, $\beta_2=0.999$, $\epsilon=10^{-8}$ \\
Schedule & 8-update linear warmup from zero, then constant learning rate \\
Regularization & Weight decay $0.1$, excluding normalization parameters; global gradient-norm clip $0.2$ \\
Auxiliary losses & No KL or entropy penalty \\
Batch size & 128 \\
Advantage & Group centering and normalization by population standard deviation; \mbox{$\epsilon=10^{-6}$} \\
Sampling & Temperature $1$, top-$p=1$; 8,192-token context; 512-token critic output cap \\
Precision & FP32 parameters, BF16 transformer computation; FP32 unembedding \\
Format reward & $-1$ for malformed/truncated feedback; no continuation for invalid feedback \\
\bottomrule
\end{tabularx}
\end{table}

\paragraph{Intervention windows.}
For both MuSiQue settings, we sample one intervention step uniformly from $\mathcal{K}=\{1,\ldots,10\}$ during training. No critique is requested if the episode terminates before the sampled step.

For ALFWorld training, the intervention step is sampled uniformly from $\mathcal{K}=\{2,\ldots,10\}$, using a task-dependent seed based on 42 and the task ID. No critique is requested if the episode has already terminated. The base-model output cap is 128 tokens per action, and the prefix and continuation share a 12-action budget. The critic is trained with one scheduled intervention and evaluated using self-call with up to three interventions and a 50-action budget.

\paragraph{Training duration and checkpoint selection.}
MuSiQue critic training stops at step 2,000, and we use this final checkpoint throughout evaluation rather than selecting the best checkpoint separately for each base model or task. This training duration was chosen because validation performance on a 20-task monitoring subset had plateaued by approximately 2,000 steps.
For ALFWorld, we similarly stop at step 800 after observing a validation plateau and use that checkpoint in all evaluations.

\section{Critic Feedback on a GRPO-Trained Generator}
\label{app:trained-generator}

\paragraph{Two-stage training.}
We first train Qwen3-4B-Instruct-2507 directly on MuSiQue in attached mode using GRPO-based optimization with the token-level DAPO objective and optimizer settings in Table~\ref{tab:training-config}, group size $G=8$, batch size 128, and maximum policy staleness four. Training uses an 8,192-token context, at least three searches, and a 128-token cap per generator response. We evaluate the retained generator checkpoint at step 1,120.

We then freeze this generator and train the critic in the attached retrieval setting using the terminal success of one generator continuation per critique ($G=8$, $N=1$). We evaluate that critic run at step 2,000. For comparison, we also pair the trained generator with the critics trained using the original instruction-tuned generator.

\paragraph{Evaluation.}
The GRPO-generator conditions use the same 200 questions and eight trials per question, temperature $1$, top-$p=1$, an 8,192-token context, at least three searches, and no total action limit. All critics use the same self-call prompt, with at most three empty \texttt{<critic/>} requests and 512 tokens per critique. The generator selects when to call; the request carries no question payload. The critic reads the visible history and supplies general feedback, which remains in the history seen by both models on subsequent turns.

\begin{table}[htbp]
\centering
\small
\setlength{\tabcolsep}{4pt}
\renewcommand{\arraystretch}{1.12}
\caption{Critic feedback on the GRPO-trained Qwen3-4B generator (step 1,120), with the original instruction-tuned generator as a reference. All evaluations with the GRPO-trained generator require at least three searches. Success rates (\%) are mean $\pm$ SEM across 200 tasks, with eight rollouts per task. Both critic rows use self-call with at most three critiques per trajectory. Bold marks the highest mean; italics mark other means whose $\pm$ SEM intervals overlap the leader's (not a significance test).}
\label{tab:trained-generator}
\begin{tabularx}{\linewidth}{Xrr}
\toprule
Critic & Attached & Wiki \\
\midrule
\multicolumn{3}{@{}l}{\textbf{Qwen3-4B before GRPO}} \\
\addlinespace[2pt]
No critic & $21.56 \pm 2.07$ & $10.38 \pm 1.59$ \\
\midrule
\multicolumn{3}{@{}l}{\textbf{Qwen3-4B after GRPO}} \\
\addlinespace[2pt]
No critic & $49.69 \pm 2.98$ & $\textit{19.69} \pm 2.36$ \\
\methodname{} trained with original generator & $51.69 \pm 2.87$ & $\mathbf{24.44} \pm 2.64$ \\
\rowcolor{criticshade}
\methodname{} trained with GRPO generator & $\mathbf{63.56} \pm 2.98$ & $\textit{23.38} \pm 2.63$ \\
\bottomrule
\end{tabularx}
\end{table}

On attached retrieval (Table~\ref{tab:trained-generator}), the critic trained with the GRPO generator improves success by $13.88$ percentage points over the no-critic condition (paired task-level SEM $2.35$ points), compared with $2.00$ points for the critic trained with the original generator. On wiki retrieval, the corresponding gains are $3.69$ points (paired task-level SEM $1.67$) and $4.75$ points (paired task-level SEM $1.83$). The GRPO-generator critic was trained only on attached retrieval; its wiki result measures transfer to a different retrieval setting. The results therefore support an additional benefit after generator training, with the largest gain in the retrieval setting used to train both models.

Our main goal is to train critics that transfer across base models and tasks, including settings where the base policy cannot be optimized directly. We do not aim to show that critic training outperforms task-specific optimization of the base policy. In this experiment, direct GRPO training yielded slightly better results than pairing the original instruction-tuned generator with a critic. However, training a critic with the GRPO-trained generator provided further gains after the generator's performance had plateaued, suggesting that the two approaches can be combined to improve task performance and achieve the better score.

\section{Results with Other Inference Protocols}
\label{app:policies}

The main results use self-call intervention, but a fixed intervention step can yield higher mean success in some settings. Fixed-step intervention is a special case of $\mathcal{P}_{\mathrm{win}}$ with $\mathcal{K}=\{k\}$. Table~\ref{tab:fixed-vs-self} gives examples from additional MuSiQue runs, comparing the same base model and critic checkpoint within each row. These diagnostic runs allow at most one self-call per trajectory, unlike the three-call limit used in the main evaluations. 

\begin{table}[htbp]
\centering
\small
\setlength{\tabcolsep}{5pt}
\renewcommand{\arraystretch}{1.12}
\caption{Selected MuSiQue comparisons where fixed-step intervention has a higher mean than self-call. Both critics are Qwen3-4B models; their training retrieval settings and base models are listed below. Success rates (\%) are mean $\pm$ SEM across 200 tasks, with eight rollouts per task. Self-call permits at most one critique. Bold marks the highest mean; italics mark other means whose $\pm$ SEM intervals overlap the leader's.}
\label{tab:fixed-vs-self}
\begin{tabular*}{\linewidth}{@{\extracolsep{\fill}}llcrr@{}}
\toprule
\multicolumn{2}{c}{Critic training} & & \multicolumn{2}{c}{Inference protocol} \\
\cmidrule(lr){1-2}\cmidrule(l){4-5}
Retrieval & Base model & $k$ & Fixed step & Self-call \\
\midrule
\multicolumn{5}{@{}l}{\textbf{Evaluation: Qwen3-4B, attached}} \\
\addlinespace[2pt]
Wiki & Qwen3-4B & $4$ & $\mathbf{28.00} \pm 2.47$ & $\textit{26.88} \pm 2.46$ \\
Attached & Qwen3.8-27B & $5$ & $\mathbf{30.50} \pm 2.32$ & $\textit{28.88} \pm 2.60$ \\
\midrule
\multicolumn{5}{@{}l}{\textbf{Evaluation: Qwen3.8-27B, wiki}} \\
\addlinespace[2pt]
Wiki & Qwen3-4B & $4$ & $\mathbf{25.94} \pm 2.63$ & $19.06 \pm 2.13$ \\
Attached & Qwen3.8-27B & $3$ & $\mathbf{25.69} \pm 2.58$ & $\textit{20.94} \pm 2.33$ \\
\bottomrule
\end{tabular*}
\end{table}

Performance can also differ substantially between neighboring intervention steps (Table~\ref{tab:neighboring-steps}). For example, with the $N=1$ critic and Qwen3-4B, moving the intervention from $k=3$ to $k=4$ raises the mean success rate from $17.75\%$ to $30.88\%$. The best reported step also depends on the base model. These are descriptive comparisons; the reported SEMs quantify uncertainty in each mean, not in the paired difference.

\begin{table}[htbp]
\centering
\footnotesize
\setlength{\tabcolsep}{3pt}
\renewcommand{\arraystretch}{1.12}
\caption{Sensitivity to neighboring intervention steps on MuSiQue-attached. Each row keeps the base model and critic checkpoint fixed. Entries are success rates (\%; mean $\pm$ SEM across 200 tasks, with eight rollouts per task). Bold marks the highest mean; italics mark other means whose $\pm$ SEM intervals overlap the leader's.}
\label{tab:neighboring-steps}
\begin{tabular*}{\linewidth}{@{\extracolsep{\fill}}lrrrr@{}}
\toprule
Base model & $k=2$ & $k=3$ & $k=4$ & $k=5$ \\
\midrule
\multicolumn{5}{@{}l}{\textbf{Terminal-outcome critic, $N=1$}} \\
\addlinespace[2pt]
Qwen3-4B & $22.12 \pm 1.92$ & $17.75 \pm 1.66$ & $\textit{30.88} \pm 2.18$ & $\mathbf{33.69} \pm 2.35$ \\
Qwen3-30B-A3B & $36.44 \pm 2.42$ & $35.44 \pm 2.19$ & $\mathbf{41.50} \pm 2.46$ & $36.38 \pm 2.29$ \\
\midrule
\multicolumn{5}{@{}l}{\textbf{Terminal-outcome critic, $N=5$}} \\
\addlinespace[2pt]
Qwen3-4B & $23.06 \pm 1.96$ & $13.81 \pm 1.37$ & $27.88 \pm 2.14$ & $\mathbf{32.56} \pm 2.30$ \\
Qwen3-30B-A3B & $\textit{34.38} \pm 2.36$ & $25.19 \pm 1.88$ & $\mathbf{35.94} \pm 2.17$ & $\textit{34.75} \pm 2.32$ \\
\bottomrule
\end{tabular*}
\end{table}

Some timing effects have a task-specific explanation. In $\tau^3$ Airline, for example, a critique at step 2 can prevent the base model from pursuing an incorrect database entry or identifier, whereas feedback at step 3 may arrive after it has already committed to that direction. Other differences are harder to explain from the trajectories. Self-call avoids choosing a fixed step for every task and base model, which motivates its use in our main evaluations.

\section{Prompts and Baseline Implementations}
\label{app:prompts}

\subsection{ALFWorld: self-call evaluation}
All three critic-assisted conditions use the same base-model system prompt, observation instruction, critic system prompt, and critic user-message template below.

\paragraph{Base model system prompt}
\begin{tcolorbox}[nebiusprompt]
\small\ttfamily\raggedright
You are an agent acting in a household text environment.\par
You can choose an environment action OR call your critic tool for advice.\par
The critic tool is a real available action, separate from the admissible environment commands. You have\par
at most 50 environment actions to finish the task. At every turn, identify the\par
earliest unfinished subgoal and take the shortest action that advances it.\par
Never repeat an action that did not change the state, revisit a location without\par
a task reason, or explore after the needed object or receptacle is known.\par
\medskip
Use these compact task recipes:\par
- place: find and take the object, go to the destination, open it if needed,\par
  then put it.\par
- clean: take the object, use the sink and faucet, retrieve it, then place it.\par
- heat: take the object, use the microwave correctly, retrieve it, then place it.\par
- cool: take the object, use the fridge correctly, retrieve it, then place it.\par
- examine under light: take the object, go to a lamp, and switch the lamp on.\par
- place two: complete both placements without undoing the first.\par
\medskip
Before every action, reason briefly in one to three short sentences about the\par
current unfinished subgoal and whether to act in the world or request a review. Keep the reasoning under\par
60 words.\par
\medskip
\medskip
You have up to THREE critic reviews. Calling the critic does not consume an environment action. The critic sees your task, environment actions, current observation, admissible actions and earlier reviews. You cannot ask it questions: use the EMPTY \textless{}critic/\textgreater{} tool action and receive general feedback. Keep useful advice in mind, check it against observations, and execute the required environment actions yourself. Every world action must match an admissible command exactly.\par
\medskip
Critic reviews are part of the normal workflow, not a last resort. Before committing to an extended search or a multi-action plan, request a review of your planned subgoals. Prefer a review over speculative exploration when the object location, required transformation, or correct completion action is unclear. Use at least one review while there is still enough action budget to apply it. After finding or taking the target object, review any uncertain remaining prerequisites before attempting completion. Reconsult when the plan stops making progress. Avoid consecutive reviews without acting on useful advice.\par
\medskip
Reply with exactly one of these two action forms, with no text outside the tags:\par
\textless{}reason\textgreater{}why an independent review is useful, at most 60 words\textless{}/reason\textgreater{}\par
\textless{}critic/\textgreater{}\par
OR\par
\textless{}reason\textgreater{}brief reasoning, at most 60 words\textless{}/reason\textgreater{}\par
\textless{}action\textgreater{}one currently admissible environment action\textless{}/action\textgreater{}\par
\end{tcolorbox}

\paragraph{Instruction appended to observations}
\begin{tcolorbox}[nebiusprompt]
\small\ttfamily\raggedright
Available tools: (1) critic review, by emitting \textless{}critic/\textgreater{}; (2) one admissible environment action.\par
Reply with exactly one of these two action forms, with no text outside the tags:\par
\textless{}reason\textgreater{}why an independent review is useful, at most 60 words\textless{}/reason\textgreater{}\par
\textless{}critic/\textgreater{}\par
OR\par
\textless{}reason\textgreater{}brief reasoning, at most 60 words\textless{}/reason\textgreater{}\par
\textless{}action\textgreater{}one currently admissible environment action\textless{}/action\textgreater{}\par
Use a review before speculative exploration or an uncertain completion. Reviews do not consume the 50 environment actions.\par
\end{tcolorbox}

\paragraph{Critic system prompt}
\begin{tcolorbox}[nebiusprompt]
\small\ttfamily\raggedright
You are a critic advising an agent acting in a household text environment.\par
Review the task, action history, current observation, and admissible actions.\par
Diagnose the trajectory so far and give concise, useful advice for what the\par
agent should consider next. Do not solve or complete the trajectory yourself.\par
Do not output environment commands, actions, or a complete action sequence.\par
Reply with exactly these two sections and no other text:\par
Analysis: \textless{}brief analysis of the trajectory so far\textgreater{}\par
Advices for next steps: \textless{}concise guidance without taking actions yourself\textgreater{}\par
\end{tcolorbox}

\paragraph{Critic user-message template}
\begin{tcolorbox}[nebiusprompt]
\small\ttfamily\raggedright
Task and initial situation:\par
\{task\}\par
\medskip
Actions taken so far:\par
\{action\_history\}\par
\medskip
Current observation:\par
\{current\_observation\}\par
\medskip
Currently admissible actions:\par
\{admissible\_actions\}\par
\medskip
Critique the trajectory so far and advise the agent how to continue.\par
\end{tcolorbox}

Each observation also states the number of critic calls remaining. After a review, the runtime supplies the current observation, admissible commands, and \texttt{Critic feedback:} followed by the parsed critique. Earlier critiques are included in subsequent critic requests under \texttt{Previous critic feedback in this trajectory (verify against the current state):}, numbered in chronological order. Once the call limit is reached, only the environment-action form is permitted.

\subsection{MuSiQue: training and self-call templates}

\paragraph{Base model prompt}
\begin{tcolorbox}[nebiusprompt]
\small\ttfamily\raggedright
You are an information-retrieval agent. Answer the multi-hop question by\par
searching hidden evidence passages. Use one action per turn. Search for\par
specific entities or relations and combine evidence across results. Do not\par
answer until you have enough evidence. You have unlimited actions.\par
A critic may be consulted after an action. Treat its feedback as advice,\par
verify it against retrieved evidence, and continue the same trajectory.\par
Reply in exactly one of these forms and nothing else:\par
\textless{}reason\textgreater{}brief reasoning, at most 60 words\textless{}/reason\textgreater{}\par
\textless{}search\textgreater{}concise search query\textless{}/search\textgreater{}\par
OR\par
\textless{}reason\textgreater{}brief reasoning, at most 60 words\textless{}/reason\textgreater{}\par
\textless{}answer\textgreater{}short final answer only\textless{}/answer\textgreater{}\par
\end{tcolorbox}

\paragraph{Self-call instruction for base model}
\begin{tcolorbox}[nebiusprompt]
\small\ttfamily\raggedright
You control whether to consult an advisor. After each search, check whether the retrieved passages support every link needed to answer the question. Consider asking for help when a bridge entity or relation is missing, passages conflict, repeated searches do not advance the evidence chain, or a proposed answer relies on memory instead of the retrieved evidence. Do not ask merely because a particular step number was reached. Verify any advice against retrieved passages before answering. Consider critic/advisor as a very important tool and consider calling it at least once: either when you're stuck and not sure what to do or verifying before answering. Critic sees the same history as you but you can't ask him any questions, he just gives general feedback. Use exactly one of these three action forms, with no text outside the tags:\par
\textless{}reason\textgreater{}brief reasoning, at most 60 words\textless{}/reason\textgreater{}\par
\textless{}search\textgreater{}concise search query\textless{}/search\textgreater{}\par
OR\par
\textless{}reason\textgreater{}brief reasoning, at most 60 words\textless{}/reason\textgreater{}\par
\textless{}answer\textgreater{}short final answer only\textless{}/answer\textgreater{}\par
OR\par
\textless{}reason\textgreater{}why advice is needed, at most 60 words\textless{}/reason\textgreater{}\par
\textless{}critic/\textgreater{}\par
\end{tcolorbox}

\paragraph{Critic system prompt}
\begin{tcolorbox}[nebiusprompt]
\small\ttfamily\raggedright
You are a concise critic advising an information-retrieval agent. Diagnose\par
whether its searches and evidence support progress on the multi-hop question,\par
and advise what information or relation to seek next. Do not answer the\par
question, do not perform a search, and do not emit \textless{}search\textgreater{} or \textless{}answer\textgreater{} tags.\par
Reply with exactly two sections and no other text:\par
Analysis: \textless{}brief diagnosis of the trajectory so far\textgreater{}\par
Advices for next steps: \textless{}concise advice about what to investigate next\textgreater{}\par
\end{tcolorbox}

\section{OOD Self-Call Analysis}
\label{app:ood-selfcall-analysis}

This appendix supports the analysis in Section~\ref{sec:analysis}.  A raw
attempt is any generated \texttt{call\_critic} action, while delivered
feedback requires a valid critic response in the trajectory.  Because the
generator chooses when to call, all comparisons are descriptive.

\subsection{How often is \methodname{} called?}

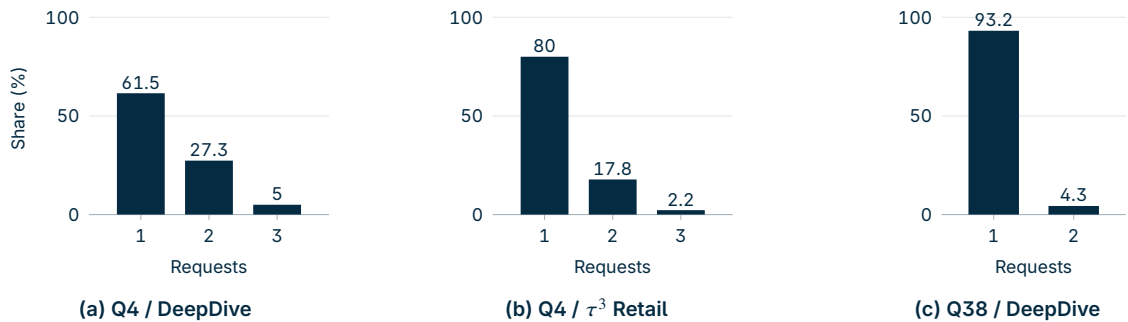
\begin{figure}[!htbp]
  \centering
  \begin{minipage}[t]{0.32\linewidth}
\centering
\begin{tikzpicture}
\begin{axis}[
  nebius axis,
  width=0.9\linewidth,height=4.4cm,ybar,bar width=18pt,ymin=0,ymax=108,
  ylabel={Share (\%)},xlabel={Requests},
  yticklabel={\pgfmathprintnumber[fixed,precision=0,assume math mode=true]{\tick}},
  symbolic x coords={1,2,3},xtick=data,
  enlarge x limits=0.38,point meta=y,
  nebius value label,
]
\addplot+[nebius single] coordinates {(1,61.491) (2,27.329) (3,4.969)};
\end{axis}
\end{tikzpicture}

\nebiussublabel{(a) Q4 / DeepDive}
\end{minipage}\hfill
\begin{minipage}[t]{0.32\linewidth}
\centering
\begin{tikzpicture}
\begin{axis}[
  nebius axis,
  width=0.9\linewidth,height=4.4cm,ybar,bar width=18pt,ymin=0,ymax=108,
  xlabel={Requests},symbolic x coords={1,2,3},xtick=data,
  yticklabel={\pgfmathprintnumber[fixed,precision=0,assume math mode=true]{\tick}},
  enlarge x limits=0.38,point meta=y,
  nebius value label,
]
\addplot+[nebius single] coordinates {(1,80.000) (2,17.778) (3,2.222)};
\end{axis}
\end{tikzpicture}

\nebiussublabel{(b) Q4 / $\tau^3$ Retail}
\end{minipage}\hfill
\begin{minipage}[t]{0.32\linewidth}
\centering
\begin{tikzpicture}
\begin{axis}[
  nebius axis,
  width=0.76\linewidth,height=4.4cm,ybar,bar width=19pt,ymin=0,ymax=108,
  xlabel={Requests},symbolic x coords={1,2},xtick=data,
  yticklabel={\pgfmathprintnumber[fixed,precision=0,assume math mode=true]{\tick}},
  enlarge x limits=0.65,point meta=y,
  nebius value label,
]
\addplot+[nebius single] coordinates {(1,93.168) (2,4.348)};
\end{axis}
\end{tikzpicture}

\nebiussublabel{(c) Q38 / DeepDive}
\end{minipage}
  \caption{Critic requests per called rollout.}
  \label{fig:ood-call-counts}
\end{figure}

Figure~\ref{fig:ood-call-counts} shows the number of requests among rollouts
that call the critic at least once.  One request dominates every setting.  The
smaller generator repeats requests more often on DeepDive; the larger
generator almost always asks once.

\begin{figure}[!htbp]
  \centering
  \begin{minipage}[t]{0.49\linewidth}
\centering
\begin{tikzpicture}
\begin{axis}[
  nebius axis,
  width=0.74\linewidth,height=4.3cm,ybar,bar width=24pt,ymin=0,ymax=84,
  ylabel={$P(\mathrm{correct})$ (\%)},xlabel={Raw attempts},
  yticklabel={\pgfmathprintnumber[fixed,precision=0,assume math mode=true]{\tick}},
  symbolic x coords={0,1,2},xtick=data,
  enlarge x limits=0.50,point meta=y,
  nebius value label,
]
\addplot+[nebius single] coordinates {(0,36.182) (1,30.303) (2,36.364)};
\end{axis}
\end{tikzpicture}

\nebiussublabel{(a) Q4 / DeepDive}
\end{minipage}\hfill
\begin{minipage}[t]{0.49\linewidth}
\centering
\begin{tikzpicture}
\begin{axis}[
  nebius axis,
  width=0.74\linewidth,height=4.3cm,ybar,bar width=24pt,ymin=0,ymax=84,
  xlabel={Raw attempts},symbolic x coords={0,1,2},xtick=data,
  yticklabel={\pgfmathprintnumber[fixed,precision=0,assume math mode=true]{\tick}},
  enlarge x limits=0.50,point meta=y,
  nebius value label,
]
\addplot+[nebius single] coordinates {(0,49.003) (1,51.333) (2,71.429)};
\end{axis}
\end{tikzpicture}

\nebiussublabel{(b) Q38 / DeepDive}
\end{minipage}
  \caption{DeepDive accuracy by the number of critic
  requests.}
  \label{fig:ood-attempt-accuracy}
\end{figure}
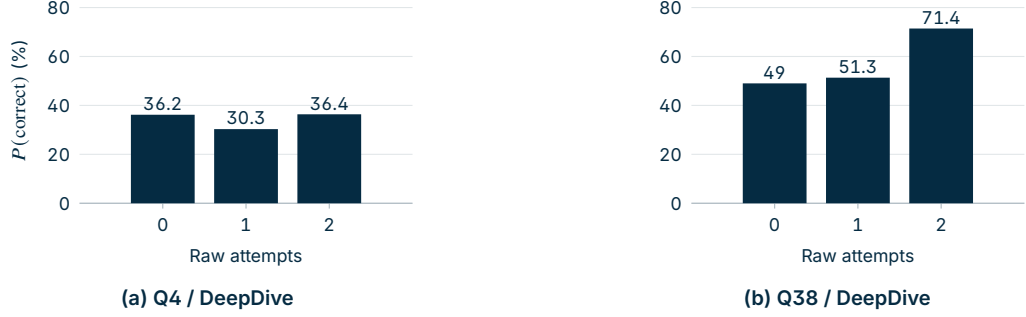

Figure~\ref{fig:ood-attempt-accuracy} relates the number of DeepDive requests
to the final answer.  A second request does not coincide with lower accuracy
for either generator.

\subsection{What advice is given, and is it used?}

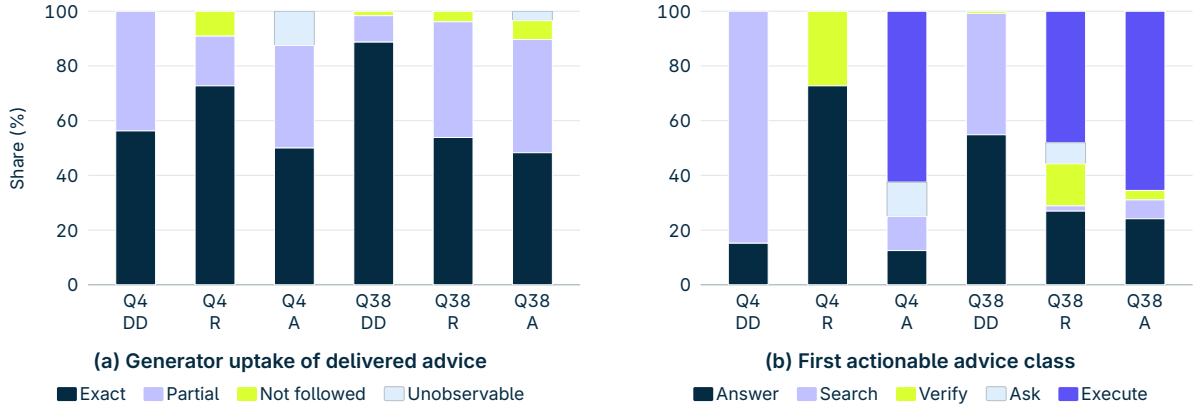
\begin{figure}[!htbp]
  \centering
  \begin{minipage}[t]{0.49\linewidth}
\centering
\begin{tikzpicture}
\begin{axis}[
  nebius axis,
  width=\linewidth,height=5.2cm,ybar stacked,bar width=15pt,ymin=0,ymax=100,
  ylabel={Share (\%)},symbolic x coords={Q4/DD,Q4/R,Q4/A,Q38/DD,Q38/R,Q38/A},xtick=data,
  yticklabel={\pgfmathprintnumber[fixed,precision=0,assume math mode=true]{\tick}},
  xticklabels={{Q4\\DD},{Q4\\R},{Q4\\A},{Q38\\DD},{Q38\\R},{Q38\\A}},
  enlarge x limits=0.12,
]
\addplot+[nebius bar=nebiuscatone] coordinates {
  (Q4/DD,56.250) (Q4/R,72.727) (Q4/A,50.000) (Q38/DD,88.710) (Q38/R,53.846) (Q38/A,48.276)
};
\addplot+[nebius bar=nebiuscattwo] coordinates {
  (Q4/DD,43.750) (Q4/R,18.182) (Q4/A,37.500) (Q38/DD,9.677) (Q38/R,42.308) (Q38/A,41.379)
};
\addplot+[nebius bar=nebiuscatthree] coordinates {
  (Q4/DD,0.000) (Q4/R,9.091) (Q4/A,0.000) (Q38/DD,1.613) (Q38/R,3.846) (Q38/A,6.897)
};
\addplot+[nebius pale bar=nebiuscatfive] coordinates {
  (Q4/DD,0.000) (Q4/R,0.000) (Q4/A,12.500) (Q38/DD,0.000) (Q38/R,0.000) (Q38/A,3.448)
};
\end{axis}
\end{tikzpicture}

\nebiussublabel{(a) Generator uptake of delivered advice}\par
{\nebiuslegendtext
\nebiuslegendswatchsmall{nebiuscatone}\,Exact\quad
\nebiuslegendswatchsmall{nebiuscattwo}\,Partial\quad
\nebiuslegendswatchsmall{nebiuscatthree}\,Not followed\quad
\nebiuslegendswatchpale{nebiuscatfive}\,Unobservable}
\end{minipage}\hfill
\begin{minipage}[t]{0.49\linewidth}
\centering
\begin{tikzpicture}
\begin{axis}[
  nebius axis,
  width=\linewidth,height=5.2cm,ybar stacked,bar width=15pt,ymin=0,ymax=100,
  symbolic x coords={Q4/DD,Q4/R,Q4/A,Q38/DD,Q38/R,Q38/A},xtick=data,
  yticklabel={\pgfmathprintnumber[fixed,precision=0,assume math mode=true]{\tick}},
  xticklabels={{Q4\\DD},{Q4\\R},{Q4\\A},{Q38\\DD},{Q38\\R},{Q38\\A}},
  enlarge x limits=0.12,
]
\addplot+[nebius bar=nebiuscatone] coordinates {
  (Q4/DD,15.179) (Q4/R,72.727) (Q4/A,12.500) (Q38/DD,54.839) (Q38/R,26.923) (Q38/A,24.138)
};
\addplot+[nebius bar=nebiuscattwo] coordinates {
  (Q4/DD,84.821) (Q4/R,0.000) (Q4/A,12.500) (Q38/DD,44.355) (Q38/R,1.923) (Q38/A,6.897)
};
\addplot+[nebius bar=nebiuscatthree] coordinates {
  (Q4/DD,0.000) (Q4/R,27.273) (Q4/A,0.000) (Q38/DD,0.806) (Q38/R,15.385) (Q38/A,3.448)
};
\addplot+[nebius pale bar=nebiuscatfive] coordinates {
  (Q4/DD,0.000) (Q4/R,0.000) (Q4/A,12.500) (Q38/DD,0.000) (Q38/R,7.692) (Q38/A,0.000)
};
\addplot+[nebius bar=nebiuscatfour] coordinates {
  (Q4/DD,0.000) (Q4/R,0.000) (Q4/A,62.500) (Q38/DD,0.000) (Q38/R,48.077) (Q38/A,65.517)
};
\end{axis}
\end{tikzpicture}

\nebiussublabel{(b) First actionable advice class}\par
{\nebiuslegendtext
\nebiuslegendswatchsmall{nebiuscatone}\,Answer\quad
\nebiuslegendswatchsmall{nebiuscattwo}\,Search\quad
\nebiuslegendswatchsmall{nebiuscatthree}\,Verify\quad
\nebiuslegendswatchpale{nebiuscatfive}\,Ask\quad
\nebiuslegendswatchsmall{nebiuscatfour}\,Execute}
\end{minipage}
  \caption{How generators use critic feedback and how the
  advice classes vary across domains.}
  \label{fig:ood-advice-distributions}
\end{figure}

Figure~\ref{fig:ood-advice-distributions} shows two complementary views.  The
generator follows most advice exactly or in part.  The content changes with
the task: DeepDive emphasizes search and answer completion, whereas $\tau^3$ more
often requires verification, clarification, or an action in the environment.

Representative examples of followed, partially followed, ignored, and
unobservable advice appear in Appendix~\ref{app:qualitative}.  They illustrate
targeted evidence gathering on DeepDive and state-aware action planning on
$\tau^3$.

Figure~\ref{fig:deepdive-advice-by-generator} in the main paper contrasts which advice is most
useful for the two frozen generators.  Targeted search helps Q4, while Q38
benefits more from advice to answer with the evidence already available.

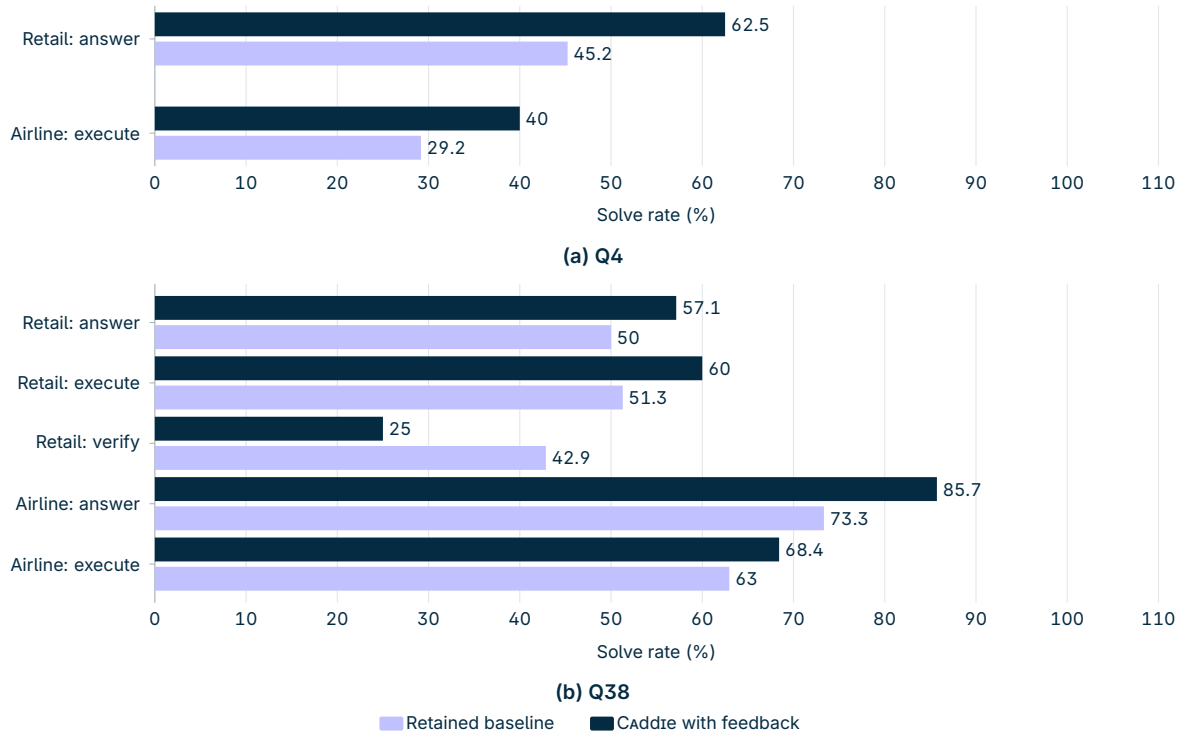
\begin{figure}[!htbp]
  \centering
  \begin{minipage}[t]{\linewidth}
\centering
\begin{tikzpicture}
\begin{axis}[
  nebius xbar axis,
  width=0.90\linewidth,height=3.7cm,xbar=2pt,bar width=9pt,xmin=0,xmax=110,
  xlabel={Solve rate (\%)},
  xticklabel={\pgfmathprintnumber[fixed,precision=0,assume math mode=true]{\tick}},
  symbolic y coords={AExecute,RAnswer},ytick=data,
  yticklabels={Airline: execute,Retail: answer},
  enlarge y limits=0.35,point meta=x,
  nodes near coords={\pgfmathprintnumber[fixed,precision=1,assume math mode=true]{\pgfplotspointmeta}},
  every node near coord/.append style={xshift=2pt,anchor=west},
]
\addplot+[nebius baseline] coordinates {(29.167,AExecute) (45.238,RAnswer)};
\addplot+[nebius caddie] coordinates {(40.000,AExecute) (62.500,RAnswer)};
\end{axis}
\end{tikzpicture}

\nebiussublabel{(a) Q4}
\end{minipage}

\vspace{2mm}
\begin{minipage}[t]{\linewidth}
\centering
\begin{tikzpicture}
\begin{axis}[
  nebius xbar axis,
  width=0.90\linewidth,height=5.8cm,xbar=2pt,bar width=9pt,xmin=0,xmax=110,
  xlabel={Solve rate (\%)},
  xticklabel={\pgfmathprintnumber[fixed,precision=0,assume math mode=true]{\tick}},
  symbolic y coords={AExecute,AAnswer,RVerify,RExecute,RAnswer},ytick=data,
  yticklabels={Airline: execute,Airline: answer,Retail: verify,Retail: execute,Retail: answer},
  enlarge y limits=0.16,point meta=x,
  nodes near coords={\pgfmathprintnumber[fixed,precision=1,assume math mode=true]{\pgfplotspointmeta}},
  every node near coord/.append style={xshift=2pt,anchor=west},
]
\addplot+[nebius baseline] coordinates {
  (62.963,AExecute) (73.333,AAnswer) (42.857,RVerify) (51.282,RExecute) (50.000,RAnswer)
};
\addplot+[nebius caddie] coordinates {
  (68.421,AExecute) (85.714,AAnswer) (25.000,RVerify) (60.000,RExecute) (57.143,RAnswer)
};
\end{axis}
\end{tikzpicture}

\nebiussublabel{(b) Q38}
\end{minipage}

\vspace{1.5mm}
\centering
{\nebiuslegendtext
\nebiuslegendswatch{baselinecolor}\,Retained baseline\qquad
\nebiuslegendswatch{caddiecolor}\,\methodname{} with feedback}
  \caption{$\tau^3$ solve rate by advice type.}
  \label{fig:ood-tau3-advice-outcomes}
\end{figure}

Figure~\ref{fig:ood-tau3-advice-outcomes} gives the $\tau^3$ counterpart to the
DeepDive comparison in Figure~\ref{fig:deepdive-advice-by-generator}.

\subsection{Complementary views}

\begin{figure}[!htbp]
\centering
\resizebox{\linewidth}{!}{\begin{minipage}{\linewidth}
\centering
\begin{minipage}[t]{0.47\linewidth}
\centering
\begin{tikzpicture}
\begin{axis}[
  nebius axis,
  width=\linewidth, height=4.8cm,
  ybar, bar width=18pt,
  ymin=-9, ymax=27, ytick={0,10,20},
  ylabel={Paired solve-rate delta (pp)},
  yticklabel={\pgfmathprintnumber[fixed,precision=0,assume math mode=true]{\tick}},
  x axis line style={draw=none},
  xtick style={draw=none},
  extra y ticks={0}, extra y tick labels={},
  extra y tick style={grid=major, grid style={draw=nebiusdeepblue!35,line width=0.4pt}},
  symbolic x coords={Scholarly,Dates,Entities},
  xtick={Scholarly,Dates,Entities},
  xticklabels={{Scholarly\\identification},{Dates /\\quantities},{Entities /\\relations}},
  enlarge x limits=0.28,
  point meta=explicit symbolic,
  nodes near coords,
  every node near coord/.append style={inner sep=1.5pt},
]
\addplot+[nebius caddie, bar shift=0pt, every node near coord/.append style={anchor=south}] coordinates {(Scholarly,20.6) [+20.6] (Entities,18.2) [+18.2]};
\addplot+[nebius baseline, bar shift=0pt, every node near coord/.append style={anchor=north}] coordinates {(Dates,-2.9) [\textendash2.9]};
\end{axis}
\end{tikzpicture}
\end{minipage}\hfill
\begin{minipage}[t]{0.47\linewidth}
\centering
\begin{tikzpicture}
\begin{axis}[
  nebius axis,
  width=\linewidth, height=4.8cm,
  ybar=2pt, bar width=16pt,
  ymin=0, ymax=33, ytick={0,10,20,30},
  yticklabel={\pgfmathprintnumber[fixed,precision=0,assume math mode=true]{\tick}},
  symbolic x coords={Commit,Targeted},
  xtick=data,
  xticklabels={{Commit to\\answer},{Targeted\\search}},
  enlarge x limits=0.45,
  point meta=y,
  nebius value label,
]
\addplot+[nebius caddie] coordinates {(Commit,27.9) (Targeted,15.1)};
\addplot+[nebius baseline] coordinates {(Commit,5.9) (Targeted,11.3)};
\end{axis}
\end{tikzpicture}
\end{minipage}

\vspace{1.5mm}
{\nebiuslegendtext
\nebiuslegendswatch{improvedcolor}\,improved\qquad
\nebiuslegendswatch{degradedcolor}\,degraded}
\end{minipage}}
\caption{Qwen3.8-27B on DeepDive with the MuSiQue-wiki \methodname{} critic. Left: paired solve-rate change (percentage points) on the tasks where the critic was called, by task family. Right: share of called rollouts that improved or degraded after the generator either committed to an answer or performed another search.}
\label{fig:deepdive-mechanisms-appendix}
\end{figure}
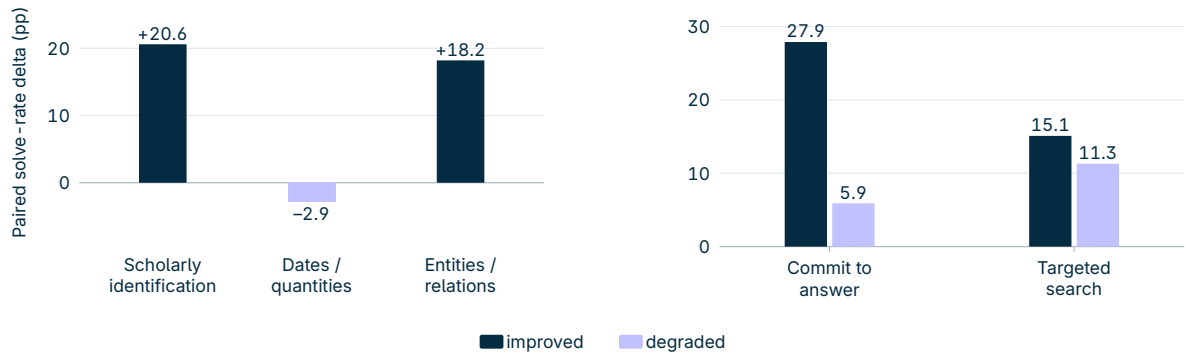

Figure~\ref{fig:deepdive-mechanisms-appendix} groups the same DeepDive behavior
by task family.  The view is consistent with the advice analysis: useful
interventions either direct the search or help the generator commit to an
answer.

\section{Performance across $\tau^3$ task families}
\label{app:tau3-clusters}

\begin{figure}[htbp]
\centering
\resizebox{\linewidth}{!}{\colorlet{clusterA}{nebiusdeepblue}
\colorlet{clusterW}{nebiusviolet}
\colorlet{clusterQ4}{nebiuslavender}
\colorlet{clusterKimi}{nebiuslime}
\colorlet{clusterRRM}{nebiuslightblueii}
\begin{minipage}{\linewidth}
\centering
\begin{tikzpicture}
\begin{groupplot}[
  group style={group size=2 by 1, horizontal sep=18pt, y descriptions at=edge left},
  nebius axis,
  scale only axis,
  height=125pt,
  ybar=1pt, /pgf/bar width=5.6pt,
  ymin=0, ymax=105, ytick={0,25,50,75,100},
  yticklabel={\pgfmathprintnumber[fixed,precision=0,assume math mode=true]{\tick}},
  xtick=data,
  enlarge x limits=0.13,
  title style={at={(0.5,1)},anchor=south,yshift=3pt,font=\nebiussemibold\footnotesize,text=nebiusdeepblue},
]
\nextgroupplot[width=227pt, ylabel={Solve rate (\%)}, title={$\tau^3$ Retail},
  symbolic x coords={Returns,Item,Order,Info},
  xticklabels={{Returns /\\refunds},{Item\\exchanges},{Order / profile\\changes},{Information\\lookup}}]
\addplot+[nebius bar=clusterA] coordinates {(Returns,67.83) (Item,51.67) (Order,51.17) (Info,78.58)};
\addplot+[nebius bar=clusterW] coordinates {(Returns,78.58) (Item,50.83) (Order,51.17) (Info,64.33)};
\addplot+[nebius bar=clusterQ4] coordinates {(Returns,64.33) (Item,46.67) (Order,42.08) (Info,73.25)};
\addplot+[nebius bar=clusterKimi] coordinates {(Returns,73.25) (Item,50.00) (Order,60.25) (Info,66.08)};
\addplot+[nebius pale bar=clusterRRM] coordinates {(Returns,76.75) (Item,54.17) (Order,52.25) (Info,66.08)};
\nextgroupplot[width=151pt, title={$\tau^3$ Airline},
  symbolic x coords={Booking,Itinerary,Cancel},
  xticklabels={{Booking /\\passenger},{Itinerary\\changes},{Cancel /\\refund}}]
\addplot+[nebius bar=clusterA] coordinates {(Booking,85.00) (Itinerary,72.50) (Cancel,72.50)};
\addplot+[nebius bar=clusterW] coordinates {(Booking,87.50) (Itinerary,75.00) (Cancel,75.00)};
\addplot+[nebius bar=clusterQ4] coordinates {(Booking,85.00) (Itinerary,71.25) (Cancel,77.50)};
\addplot+[nebius bar=clusterKimi] coordinates {(Booking,52.50) (Itinerary,66.25) (Cancel,65.00)};
\addplot+[nebius pale bar=clusterRRM] coordinates {(Booking,70.00) (Itinerary,67.50) (Cancel,70.00)};
\end{groupplot}
\end{tikzpicture}

\vspace{1.5mm}
{\nebiuslegendtext
\nebiuslegendswatch{clusterA}\,\methodname{}-A\quad
\nebiuslegendswatch{clusterW}\,\methodname{}-W\quad
\nebiuslegendswatch{clusterQ4}\,Prompted Q4\quad
\nebiuslegendswatch{clusterKimi}\,Kimi K3\quad
\nebiuslegendswatchpale{clusterRRM}\,Agent-RRM}
\end{minipage}}
\caption{Qwen3.8-27B solve rate by $\tau^3$ task family. Each
group uses the same tasks and replicates for every critic; A/W denote the
MuSiQue-attached/wiki \methodname{} checkpoints.}
\label{fig:tau3-intent-clusters}
\end{figure}
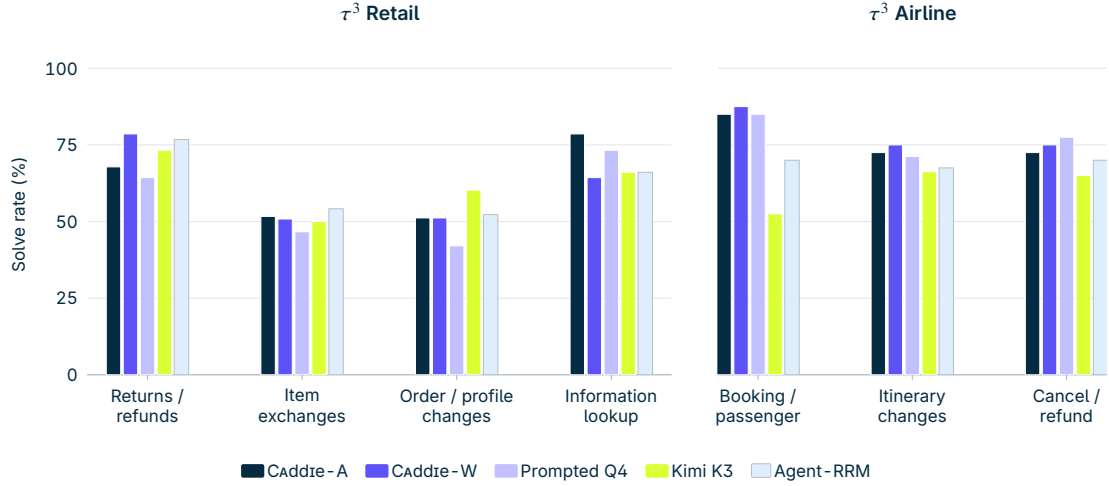

We cluster $\tau^3$ tasks by the user's primary intent and aggregate matched
replicates within each family (Figure~\ref{fig:tau3-intent-clusters}).  This analysis compares \methodname{} with the
prompted critics and Agent-RRM on identical task subsets, revealing which
interaction types account for differences in aggregate performance.

\raggedbottom
\section{Qualitative OOD Trajectories}
\label{app:qualitative}

We show representative task-matched trajectories from the analysis
in Section~\ref{sec:analysis}. We compare advice from \methodname{} with the
critic baselines, examining how each intervention changes the generator's
behavior and subsequent trajectory. The ``Advice'' rows reproduce the stored
advice verbatim; the surrounding ``Before'' and ``After'' rows summarize the
trajectory for readability.

\newcommand{\originaladvice}[1]{{\fontsize{6.8}{7.4}\selectfont\itshape #1}}
\newcommand{\compactexample}{%
  \setlength{\tabcolsep}{2.5pt}%
  \renewcommand{\arraystretch}{0.94}%
}
\tcbset{qualexample/.style={boxsep=1.5pt,left=3pt,right=3pt,
  top=2pt,bottom=2pt,before skip=3pt,after skip=3pt}}

\subsection{Preserving transactional state on $\tau^3$ Retail}

\begin{tcolorbox}[qualexample,colback=analysisfill,colframe=analysisblue!75,
  coltext=newblue,boxrule=0.5pt,arc=1.5pt,breakable]
\footnotesize\raggedright\compactexample
\textbf{Original payment method.} The user asks to return a damaged bicycle to
the card used for its order.\par
\begin{tabularx}{\linewidth}{@{}>{\bfseries\raggedright\arraybackslash}p{0.10\linewidth}>{\raggedright\arraybackslash}X>{\raggedright\arraybackslash}X@{}}
 & \textbf{Prompted Q4} & \textbf{\methodname{} (attached)} \\
Before & Finds the bicycle order, whose payment history names Mastercard ending in 8484. & Finds the same order, item, and original Mastercard. \\
Advice & \originaladvice{Proceed with issuing the refund to the Visa card ending in 6791 using the confirmed order and item details. Confirm the refund amount, tracking ID, and estimated processing time to the user before concluding the conversation.} & \originaladvice{Call the request\_refund function with the item\_id ``7758198585'' and payment\_method ``credit\_card\_8105988''.} \\
After & Uses the wrong card; the return is rejected and the trajectory transfers to a human. & Returns the bicycle to the original payment method; the task succeeds. \\
\end{tabularx}
\end{tcolorbox}

\begin{tcolorbox}[qualexample,colback=analysisfill,colframe=analysisteal!80!black,
  coltext=newblue,boxrule=0.5pt,arc=1.5pt,breakable]
\footnotesize\raggedright\compactexample
\textbf{Ordering irreversible actions.} The user asks to exchange hiking boots
and return two other items from the same delivered order, prioritizing the
exchange.\par
\begin{tabularx}{\linewidth}{@{}>{\bfseries\raggedright\arraybackslash}p{0.10\linewidth}>{\raggedright\arraybackslash}X>{\raggedright\arraybackslash}X@{}}
 & \textbf{Kimi K3} & \textbf{\methodname{} (wiki)} \\
Before & Has the correct order, replacement boots, return items, and payment method. & Has the same state and the user's explicit request to complete the exchange first. \\
Advice & \originaladvice{Process the return for the hose and backpack and the exchange for the hiking boots to the confirmed waterproof variant, applying all refunds (item refund and \$23.99 difference) to the original PayPal account. Then clearly confirm completion to the user.} & \originaladvice{Proceed with the exchange using the original order's payment method (PayPal) as it was already linked to the user's account. Provide a clear, concise confirmation message like: ``Your hiking boots will be exchanged to the waterproof version. A \$23.99 refund will be issued to your original PayPal account. Is this correct?'' Then wait for the user to reply with ``yes'' to complete the exchange request.} \\
After & Submits the return first, changing the order state and making the boot exchange unavailable. & Completes the boot exchange first, satisfying the user's prioritized action. \\
\end{tabularx}
\end{tcolorbox}

\begin{tcolorbox}[qualexample,colback=analysisfill,colframe=neutralcolor!85!black,
  coltext=newblue,boxrule=0.5pt,arc=1.5pt,breakable]
\footnotesize\raggedright\compactexample
\textbf{Tracking a changed request.} During a two-item exchange, the user
changes the request and ultimately asks only to return a water bottle.\par
\begin{tabularx}{\linewidth}{@{}>{\bfseries\raggedright\arraybackslash}p{0.10\linewidth}>{\raggedright\arraybackslash}X>{\raggedright\arraybackslash}X@{}}
 & \textbf{Agent-RRM} & \textbf{\methodname{} (attached)} \\
Before & Reviews the earlier exchange plan before the user's final correction. & Requests feedback after the user has withdrawn both exchanges and confirmed the return. \\
Advice & \originaladvice{\textbullet\ Correctly used find\_user\_id\_by\_name\_zip to verify identity and obtain a user ID.\newline
\textbullet\ Properly called get\_user\_details to retrieve full account information.\newline
\textbullet\ Accurately identified the order containing both items with get\_order\_details.\newline
\textbullet\ Appropriately called get\_product\_details for each product to find matching variants.\newline
\textbullet\ Filtered variants based on user preferences and presented a clear list.\newline
\textbullet\ Handled the unavailable original lamp variant correctly.\newline
\textbullet\ Minor: did not explicitly ask the user to confirm return or keep items, but this does not impact tool usage.\newline
\textbullet\ No irrelevant, repeated, or malformed tool calls; no fabricated arguments or filenames.} & \originaladvice{Call the request\_return function with the parameter item\_id set to ``8538875209''.} \\
After & Executes a lamp exchange; the later return cannot be completed and the trajectory transfers. & Returns only the water bottle to the original PayPal method; the task succeeds. \\
\end{tabularx}
\end{tcolorbox}

\begin{tcolorbox}[qualexample,colback=analysisfill,colframe=analysisgold!85!black,
  coltext=newblue,boxrule=0.5pt,arc=1.5pt,breakable]
\footnotesize\raggedright\compactexample
\textbf{Ordering dependent updates.} The user asks to change both the shipping
address and an item in the same pending order; either item modification locks
the address.
\begin{tabularx}{\linewidth}{@{}>{\bfseries\raggedright\arraybackslash}p{0.10\linewidth}>{\raggedright\arraybackslash}X>{\raggedright\arraybackslash}X@{}}
 & \textbf{Prompted Q4} & \textbf{\methodname{} (wiki)} \\
Before & Has the prior shipping address, replacement watch, and confirmation for both changes. & Has the same address, item identifiers, payment method, and confirmation. \\
Advice & \originaladvice{Verify the shipping address by retrieving it from the order \#W3445693 (sneakers and air purifier) since the user specified it is the same as that order. Then confirm whether the user wants to change the silicone-to-metal strap on the white-dial variant, noting the price difference and ensuring the change is eligible---especially since the blue-dial metal version is unavailable. Only after confirming these details should the next update be initiated.} & \originaladvice{Call the update\_order function with the following parameters:\newline
\textbullet\ order\_id: ``\#W4219264''\newline
\textbullet\ new\_shipping\_address: ``144 Lakeview Drive, Suite 925, New York, NY 10228, USA''\newline
\textbullet\ product\_variant: \{``item\_id'': ``2407258246'', ``strap material'': ``metal'', ``dial color'': ``white''\}\newline
After successful update, respond to the user with: ``Your order has been successfully updated with the new shipping address and the metal watch with a white dial. Thank you!''} \\
After & Modifies the item first; the address is then locked and requires a human transfer. & Updates the address first and then the watch; both requested changes succeed. \\
\end{tabularx}
\end{tcolorbox}

\begin{tcolorbox}[qualexample,colback=analysisfill,colframe=analysismauve!80!black,
  coltext=newblue,boxrule=0.5pt,arc=1.5pt,breakable]
\footnotesize\raggedright\compactexample
\textbf{Following the latest intent.} After exploring a camera upgrade, the
user abandons the modification and asks to cancel the pending order.
\begin{tabularx}{\linewidth}{@{}>{\bfseries\raggedright\arraybackslash}p{0.10\linewidth}>{\raggedright\arraybackslash}X>{\raggedright\arraybackslash}X@{}}
 & \textbf{Agent-RRM} & \textbf{\methodname{} (attached)} \\
Before & Reviews the earlier camera-variant plan after the user has switched to cancellation. & Sees the same revision and the still-pending order. \\
Advice & \originaladvice{\textbullet\ Tool calls are well sequenced: email $\rightarrow$ user $\rightarrow$ orders $\rightarrow$ order details $\rightarrow$ product variants $\rightarrow$ price difference. No unnecessary or irrelevant calls.\newline
\textbullet\ The agent correctly extracted the zoom level from the order and informed the user.\newline
\textbullet\ When the user asked for a 10x zoom, the agent used get\_product\_details to find the exact variant, which is appropriate.\newline
\textbullet\ The agent calculated the price difference correctly.\newline
\textbullet\ Minor: the agent did not explicitly confirm that the user still wants the same resolution and storage before proceeding, but this is implied by the order data and does not affect tool usage.\newline
\textbullet\ No hallucination of tool arguments, filenames, or outputs; all tool calls are grounded in the provided data.} & \originaladvice{Call the cancel\_order function with order\_id ``\#W9284598'' and cancel\_reason ``ordered by mistake''.} \\
After & Transfers the still-actionable cancellation to a human. & Confirms and cancels the order, refunding the original Visa; the task succeeds. \\
\end{tabularx}
\end{tcolorbox}

\subsection{Preserving itinerary constraints on $\tau^3$ Airline}

\begin{tcolorbox}[qualexample,colback=analysisfill,colframe=analysisteal!80!black,
  coltext=newblue,boxrule=0.5pt,arc=1.5pt,breakable]
\footnotesize\raggedright\compactexample
\textbf{Global rather than local cheapest route.} The user requests the
cheapest economy itinerary from Atlanta to Philadelphia on the following day.
\begin{tabularx}{\linewidth}{@{}>{\bfseries\raggedright\arraybackslash}p{0.10\linewidth}>{\raggedright\arraybackslash}X>{\raggedright\arraybackslash}X@{}}
 & \textbf{Kimi K3} & \textbf{\methodname{} (wiki)} \\
Before & Retains the original connection through Chicago and prices that route at \$216. & Compares the available itineraries and finds a \$207 connection through New York. \\
Advice & \originaladvice{Execute the reservation change immediately using update\_reservation\_flights with reservation\_id M05KNL, cabin ``economy'', the two confirmed flights, and payment\_id ``gift\_card\_8887175'' since the user explicitly requested the original payment method. Do not ask for further confirmation---all details have been validated and the user has given explicit consent. After the change completes, calculate the refund amount (\$2,787 $-$ \$216 $=$ \$2,571) and confirm the refund will be processed to the original gift card.} & \originaladvice{The final action should be a simple, clear confirmation of the flight change. Since the user has confirmed the change and the refund to the original payment method, the next decision should be to generate a final response confirming the updated travel details and the refund amount. For example: ``Your flight has been successfully changed to May 24. Your new route is ATL $\rightarrow$ LGA $\rightarrow$ PHL with a total cost of \$207. A refund of \$2,580 will be issued to your original payment method (gift card). Thank you for choosing our service!'' Then end the conversation.} \\
After & Changes the reservation to the more expensive route. & Books the cheapest itinerary and refunds the difference to the original gift card; the task succeeds. \\
\end{tabularx}
\end{tcolorbox}

\begin{tcolorbox}[qualexample,colback=analysisfill,colframe=analysisblue!75,
  coltext=newblue,boxrule=0.5pt,arc=1.5pt,breakable]
\footnotesize\raggedright\compactexample
\textbf{Completing a two-stage fare change.} A Basic Economy reservation must
first be upgraded before it can be moved to a cheaper nonstop flight.
\begin{tabularx}{\linewidth}{@{}>{\bfseries\raggedright\arraybackslash}p{0.10\linewidth}>{\raggedright\arraybackslash}X>{\raggedright\arraybackslash}X@{}}
 & \textbf{Prompted Q4} & \textbf{\methodname{} (wiki)} \\
Before & Finds the required upgrade, its temporary charge, and the cheaper nonstop itinerary. & Establishes the same policy constraint, payment method, and nonstop option. \\
Advice & \originaladvice{Verify with the user whether they are willing to proceed with the economy upgrade and confirm that the requested nonstop change is possible under current policy. Then, validate the total cost of the upgrade and flight change using accurate pricing data before finalizing the action. Avoid speculative calculations without user confirmation or system validation.} & \originaladvice{Call the correct function: upgrade\_reservation with the reservation ID ``OWZ4XL'' and new cabin class ``economy''. Set the payment method to ``credit\_card\_9659780'' (Visa ending 1777). After successfully processing this action, confirm the upgrade and return a clear message that it has been completed. Then, follow up with a separate step to change the entire flight to the nonstop option (HAT041 from EWR to LAX).} \\
After & Treats the intermediate upgrade charge as exceeding the budget and transfers the user. & Executes both steps; the second change more than offsets the upgrade and the requested itinerary succeeds. \\
\end{tabularx}
\end{tcolorbox}

\subsection{Committing to retrieved evidence on DeepDive}

\begin{tcolorbox}[qualexample,colback=analysisfill,colframe=analysisgold!85!black,
  coltext=newblue,boxrule=0.5pt,arc=1.5pt,breakable]
\footnotesize\raggedright\compactexample
\textbf{IEEE service-chain paper.} The task asks for the exact title of a 2019
paper on reliability-guaranteed deployment of service-function chains.\par
\begin{tabularx}{\linewidth}{@{}>{\bfseries\raggedright\arraybackslash}p{0.10\linewidth}>{\raggedright\arraybackslash}X>{\raggedright\arraybackslash}X@{}}
 & \textbf{Prompted Q4} & \textbf{\methodname{} (wiki)} \\
Before & Retrieves the correct author, venue, DOI, and candidate title. & Retrieves the same decisive bibliographic evidence. \\
Advice & \originaladvice{Verify the Stanford affiliation of the textbook co-author and cross-check if any of Young-Han Kim's 2019 papers on service chains or redundancy include a co-author from a Seoul-based university. Specifically, search for ``Young-Han Kim 2019 IEEE paper service chain redundancy Seoul co-author'' to confirm the required collaboration and match the full profile. Also, investigate if any other prominent information theorist with Stanford ties (e.g., in network information theory) meets the criteria to rule out the primary candidate.} & \originaladvice{Return the final answer: ``An Efficient Reliability Guaranteed Deployment Scheme for Service Function Chains''. This is the exact title of the intended paper and meets all the criteria: it is published in an IEEE-accessible venue (IEEE Access), fits the described research topic, and includes the correct author profile.} \\
After & Spends the remaining budget investigating the false mismatch and submits no valid title. & Returns \emph{An Efficient Reliability Guaranteed Deployment Scheme for Service Function Chains}; the task succeeds. \\
\end{tabularx}
\end{tcolorbox}

\begin{tcolorbox}[qualexample,colback=analysisfill,colframe=analysismauve!80!black,
  coltext=newblue,boxrule=0.5pt,arc=1.5pt,breakable]
\footnotesize\raggedright\compactexample
\textbf{Semiconductor growth paper.} The task asks for the exact title of a
paper on ex-situ calibration for commercial semiconductor heterostructures.
\begin{tabularx}{\linewidth}{@{}>{\bfseries\raggedright\arraybackslash}p{0.10\linewidth}>{\raggedright\arraybackslash}X>{\raggedright\arraybackslash}X@{}}
 & \textbf{Prompted Q4} & \textbf{\methodname{} (wiki)} \\
Before & Follows a different X-ray characterization paper whose authors and technical description do not match the clues. & Retrieves the exact title, abstract language, journal, year, and author affiliations. \\
Advice & \originaladvice{Investigate the full authorship and affiliations of the 2003 ``Characterization of MBE grown II-VI semiconductor thin layers by X-ray interference'' paper to confirm links to Surrey or Essex institutions with fellowships. Cross-reference Brian Cavenett's biography and publications to establish his PhD timeline, THz research, and industrial collaborations (especially in automotive). Then, search for publications from that period jointly citing ``XRD,'' ``calibration,'' and ``MBE'' with a co-author from Surrey or Essex who is a Fellow of the IET or FRSE.} & \originaladvice{Provide the final answer: ``Precision growth for the manufacture of semiconductor heterostructure devices''. This is the correct and most relevant paper title based on all the described conditions.} \\
After & Continues around the false lead and does not return the required title. & Returns \emph{Precision growth for the manufacture of semiconductor heterostructure devices}; the task succeeds. \\
\end{tabularx}
\end{tcolorbox}

{\color{newblue}
Across these cases, \methodname{} preserves the decisive state needed for the
next action.  The comparator advice instead substitutes an unseen value, drops
operation order, narrows the candidate set too early, or summarizes a stale
request.
}
\flushbottom

\section{Additional Discussion and Limitations}
\label{app:limitations}

\paragraph{Task structure.}
In our MuSiQue experiments, the base model gathers missing evidence through retrieval calls. However, in self-contained tasks, such as mathematical reasoning, the critic may instead have enough information to produce a complete solution in a single step. In our experiments on step-by-step mathematical reasoning, the critic increasingly supplied solutions that the base model could follow directly, with this behavior emerging after a few hundred training steps.

The training signal can favor this behavior. Consider a group of $G$ critiques, each evaluated using $N>1$ base-model continuations. General advice from the critic can steer the base model toward a particular solution, but the base model must still work out the remaining steps, so the continuations may have different outcomes and yield an intermediate average reward. If the base model reliably follows a complete solution given by the critic, the continuations tend to agree on the outcome. Correct solutions then receive rewards near one, and incorrect solutions receive rewards near zero, concentrating their rewards near the two endpoints.

Within a mixed group, complete solutions that succeed in every continuation receive the largest positive advantages. For such a solution and another critique with $R_i<1$, the group-normalized advantage in Section~\ref{sec:method:optimization} gives $\hat A_{\mathrm{sol}}-\hat A_i=(1-R_i)/\sigma_R>0$, where $\sigma_R>0$ is the group's reward standard deviation. A consistently failing solution instead receives a negative advantage whenever the group contains successful continuations. Repeated updates can therefore reinforce correct solution generation and suppress incorrect solutions, effectively teaching the critic to solve the task itself. This incentive depends on higher mean rewards, not on reduced variance alone.

Direct solution generation can also be optimal under the critic's objective. With binary task rewards, a correct solution that the base model always copies achieves expected reward one, the maximum possible, and cannot be outperformed by advice that leaves a chance of failure. If the critic always provides a complete solution and the base model reproduces it, the critic's expected task reward reduces to
\begin{equation}
  J(\theta)=
  \mathbb{E}_{s_k\sim d_{\mathrm{win}},\;c\sim\pi_c^{\theta}(\cdot\mid s_k)}
  \bigl[r_{\mathrm{sol}}(s_k,c)\bigr],
  \label{eq:critic-as-solver}
\end{equation}
where $d_{\mathrm{win}}$ is the distribution of sampled state prefixes and $r_{\mathrm{sol}}(s_k,c)$ is the task reward for the solution contained in $c$. In this limiting case, the critic is optimized as a solver, while the base model's parameters remain fixed. This can improve task success without establishing that the critic has learned guidance that transfers across base models or tasks. Tasks that require further environment observations make this strategy less applicable, but tool use alone does not rule it out.

Advisor Models \citep{advisor2026} also study a related self-contained mathematical reasoning setting. Their released implementation generates advice for a completed initial attempt and rewards the correctness of a single revised base-model response, without further tool interaction. They report 20 training epochs, but do not analyze whether the advisor increasingly supplies complete solutions. Training duration may influence when this behavior appears. 

\paragraph{Continuation sampling.}
A critique's reward depends on how the base model responds to it. For binary task rewards, let $q_i$ denote the success probability of a continuation from $s_k\oplus c^{(i)}$. If the $N$ continuations are conditionally independent, the reward estimate satisfies
\begin{equation}
  \operatorname{Var}\!\left(R_i\mid s_k,c^{(i)}\right)
  =\frac{q_i(1-q_i)}{N}.
  \label{eq:continuation-variance}
\end{equation}
Thus, critiques that lead to nearly certain success or failure have lower conditional reward variance. Increasing $N$ reduces estimation noise, but does not necessarily accelerate convergence. In our experiments, $N=5$ yielded validation curves similar to $N=1$, with smaller fluctuations later in training and similar final performance (Appendix~\ref{app:reward_variants}), at the cost of additional base-model continuations.

\paragraph{Environment restoration.}
Training requires continuing multiple branches from the same intermediate state. For retrieval over a fixed corpus, this can often be done by replaying the trajectory prefix. In stateful environments, the external environment must also be restored. For example, software-engineering tasks may require checkpointing files and running processes. Restoring equivalent states across branches can therefore be difficult or costly, particularly when tools interact with external services.

\paragraph{Critic scale.}
We train only Qwen3-4B critics. Although we evaluate their transfer to larger base models, we do not establish how increasing the critic's capacity affects training, performance, or transfer.

\paragraph{Training cost.}
Reaching a performance plateau in our experiments required approximately 2,000 training steps. Each sampled state requires $GN$ base-model continuations, so training cost depends on the base model and environment as well as the critic. Expensive tool execution, long trajectories, and larger base models can substantially increase CPU and GPU requirements.

\end{document}